\documentclass[10pt,twocolumn,letterpaper]{article}

\usepackage{cvpr}              
\usepackage{multicol}
\usepackage{multirow}
\usepackage{adjustbox}
\usepackage{subcaption}
\usepackage[dvipsnames]{xcolor}
\usepackage[dvipsnames]{xcolor}

\definecolor{cvprblue}{rgb}{0.21,0.49,0.74}
\definecolor{comment}{rgb}{0.8,0.2,0.1}

\usepackage[pagebackref,breaklinks,colorlinks,allcolors=cvprblue]{hyperref}

\def\paperID{2032} 
\def\confName{CVPR}
\def\confYear{2024}

\title{Balancing Act:  Distribution-Guided Debiasing in Diffusion Models}

\author{%
  \begin{tabular}[t]{c}
    Rishubh Parihar\thanks{Equal contribution} $^{,1}$ \quad
    Abhijnya Bhat$^{*,1}$  \quad
    Abhipsa Basu$^1$   \quad 
    Saswat Mallick$^1$ \\ 
    Jogendra Nath Kundu$^2$ \quad 
    R. Venkatesh Babu$^1$ \\
    \end{tabular}%
  \quad
  \and
  \begin{tabular}[t]{c}
    $^1$Indian Institute of Science, Bangalore \quad 
    $^2$Meta Reality Labs \\
  \end{tabular}%
}

\begin{document}
\maketitle
\begin{abstract}
Diffusion Models (DMs) have emerged as powerful generative models with unprecedented image generation capability. These models are widely used for data augmentation and creative applications. However, DMs reflect the biases present in the training datasets. This is especially concerning in the context of faces, where the DM prefers one demographic subgroup vs others (eg. female vs male). 
In this work, we present a method for debiasing DMs without relying on additional reference data or model retraining. Specifically, we propose \textbf{Distribution Guidance}, which enforces the generated images to follow the \underline{prescribed attribute distribution}. To realize this, we build on the key insight that the latent features of denoising UNet hold rich demographic semantics, and the same can be leveraged to guide debiased generation.
We train \textbf{Attribute Distribution Predictor} (ADP) - a small mlp that maps the latent features to the distribution of attributes. ADP is trained with pseudo labels generated from existing attribute classifiers. The proposed Distribution Guidance with ADP enables us to do fair generation.
Our method reduces bias across single/multiple attributes and outperforms the baseline by a significant margin for unconditional and text-conditional diffusion models. Further, we present a downstream task of training a fair attribute classifier by augmenting the training set with our generated data.

\noindent Code is available at - \href{https://ab-34.github.io/balancing_act/}{project page}.

\end{abstract}    

\let\origaddcontentsline\addcontentsline
\renewcommand{\addcontentsline}[3]{} 

\vspace{-4mm}
\section{Introduction}
\vspace{-2mm}
\begin{figure}[t]
    \centering
    \includegraphics[width=0.9\linewidth]{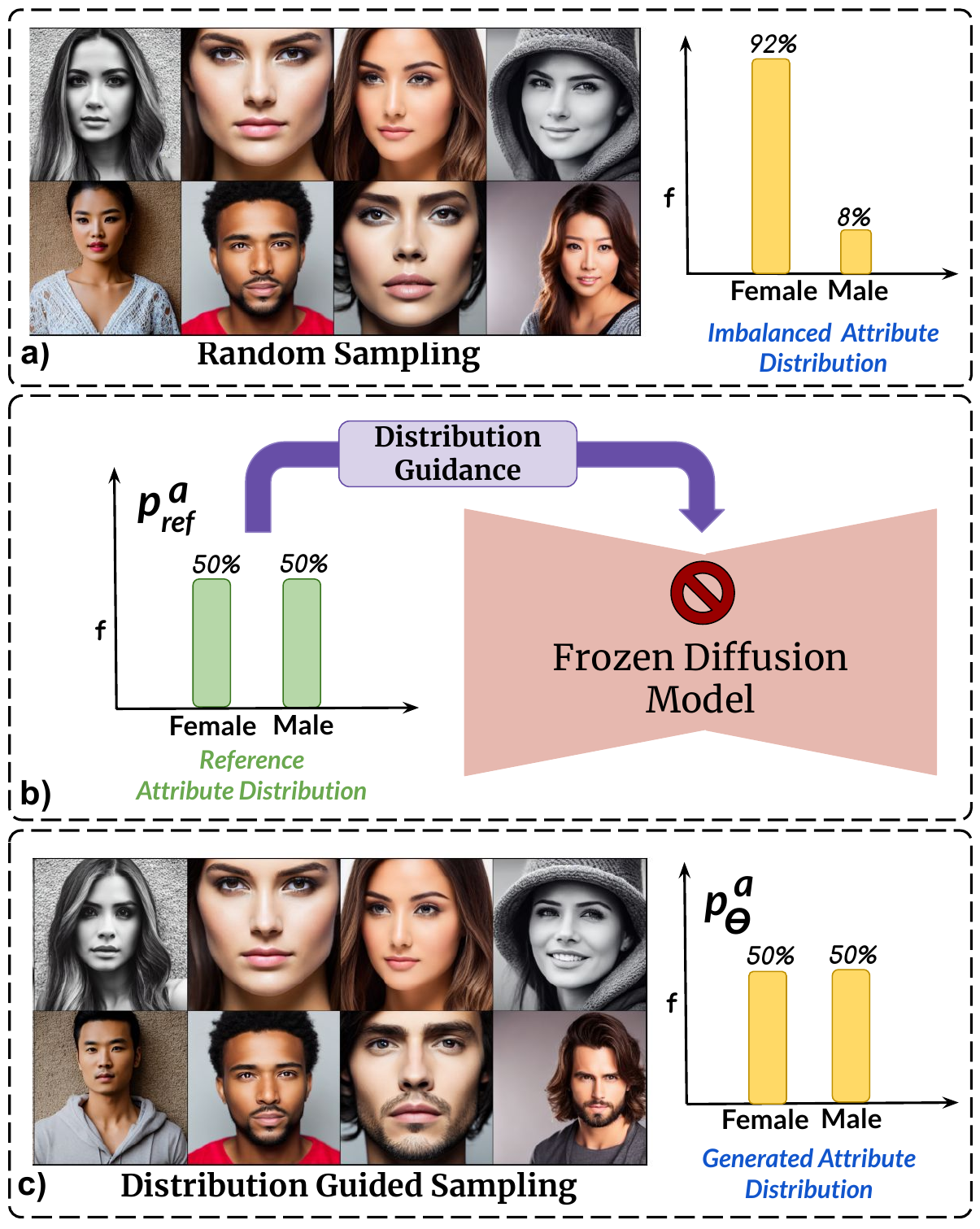}
    \vspace{-3mm}
     \caption{\textbf{a)} Random sampling from a pretrained DM~\cite{ldm} generates images with gender imbalance. \textbf{b)} Proposed method takes a user-defined reference attribute $\mathbf{p^a_{ref}}$ and performs \textit{distribution guidance} on a pretrained DM. \textbf{c)} Sampling with distribution guidance results in \textit{fair} generation that follow user define $\mathbf{p^a_{ref}}$.}
     \vspace{-8mm}
    \label{fig:teaser}
\end{figure}

Recent advancements in Diffusion Models (DM)~\cite{ddpm, ldm, ddim}
have garnered much interest among researchers in evaluating the quality of the generated content. These models are not only used to generate realistic content but also to augment real datasets~\cite{diffusion_augms, melzi2023gandiffface}
for downstream tasks. However, existing DMs have been found to synthesize biased content with respect to multiple demographic factors like gender, race, \etc, which can be detrimental to society once these models are deployed in the real world~\cite{perera2023analyzing, luccioni2023stable, rosenberg2023unbiased}. The problem is largely caused by the images used to train these models, as the outputs of these models are governed by these training datasets~\cite{perera2023analyzing, maluleke2022studying}. 
Effects of such harmful biases have been shown by multiple recent works involving studies on DMs~\cite{perera2023analyzing, luccioni2023stable, rosenberg2023unbiased}, GANs and other generative models~\cite{maluleke2022studying, choi2020fair, humayun2021magnet}. In fact, Perera et al.~\cite{perera2023analyzing} show that unconditional DMs--even when trained with balanced data--amplify racial biases, leading to the generation of more white-skinned faces than dark-skinned ones. Biases in the generated data are even more evident in large text-to-image DMs, e.g., models mostly tend to generate a specific gender with a given profession (like male and doctor)~\cite{mandal2023measuring, zhang2023iti, t2i-debiasing-safe-ldm}. 



Existing works on debiasing, either require a reference dataset ~\cite{choi2020fair,zhang2023iti} and/or allow retraining of the model~\cite{choi2020fair,xu2018fairgan,yu2020inclusive}. On the contrary, our work aims to mitigate biases in both unconditional and conditional DMs, enabling \textit{fair} generation without model retraining. We propose a practical setting where we provide a \textit{reference attribute distribution} ($\mathbf{p^a_{ref}}$) for sensitive attribute $a$ and query the DM to sample images following $\mathbf{p^a_{ref}}$. This framework allows for \textit{adapting} any existing DM to a pre-defined distribution. \Eg, a user can provide $\mathbf{p^a_{ref}}$ as a uniform distribution for a given sensitive attribute to generate \textit{balanced} attribute distribution. We believe defining $\mathbf{p^a_{ref}}$ provides just enough information to condition the DM for \textit{fair} generation. This is an extremely practical setting for debiasing and is particularly important for large text-to-image DMs~\cite{ldm, imagen,dalle} where retraining or fine-tuning is computationally intensive. 


One plausible approach for \textit{fair} generation is to guide every generated sample with attribute classifiers following classifier guidance~\cite{clf_ddpm}.
However, such a framework, though simple, is overly restrictive as it requires 
\textit{presetting} and enforcing attributes for each sample individually (which we call sample guidance). Such constraints during denoising result in inferior generation quality, as discussed in Sec.4.1. Instead, we propose to \textit{jointly} denoise a whole batch of samples and guide the process with $\mathbf{p^a_{ref}}$ (which we call distribution guidance). Specifically, we push the generated batch attribute distribution $\mathbf{{p^a_\theta}}$ and  $\mathbf{p^a_{ref}}$ close to each other during the reverse process. Distribution guidance provides more flexibility to each sample during generation as it does not enforce a preset of attributes on a sample basis. Intuitively, distribution guidance prioritizes transforming easier samples close to the decision boundary. This results in \textit{fair} generation without sacrificing the generation quality.  

A major challenge for guidance-based conditioning is that it requires separate image classifiers for each noise scale of the diffusion process. To overcome this, we propose to perform guidance in a semantically rich feature space - \textit{h-space}~\cite{h_space} of DMs. Specifically, we train an \textbf{Attribute Distribution Predictor (ADP)} that predicts attribute distribution directly from the \textit{h-space} features. As ADP is trained on rich and discriminative \textit{h-space} features, it - \textit{a) is implemented as a linear layer, b) requires minimal training data, and c) is fast in training and inference}. Finally, during inference, we steer the h-space representation by matching the predictions from ADP to $\mathbf{p^a_{ref}}$.

We extensively evaluate our proposed method for the fair generation of single and multi-attribute cases for face generation. Additionally, we present the results of our method on Stable Diffusion~\cite{ldm}, a large text-to-image DM. Further, as downstream application train \textit{debiased} attribute classifiers by augmenting the training data for minority subgroups.


The major contributions of this work are the following:
\begin{enumerate}
    \item A novel setting for debiasing existing DMs without retraining, given a reference attribute distribution.
    \item Distribution guidance to condition the reverse diffusion process on a reference attribute distribution.
    \item Propose guidance in the intermediate features of diffusion network (\textit{h-space}), which leads to data-efficient training and fast generation.
\end{enumerate}
\label{sec:intro}
\vspace{-4mm}

\section{Related Works}
\vspace{-2mm}

\noindent \textbf{Biases in Generative Models}. While generative models like Generative Adversarial Networks and Diffusion Models have become the de-facto tools for image generation in recent times, studies show that they are not free of biases \cite{perera2023analyzing, luccioni2023stable, rosenberg2023unbiased, maluleke2022studying, jain2020imperfect, basu2023inspecting}. Perera et al. ~\cite{perera2023analyzing} show that unconditional diffusion models amplify the biases in the training data with respect to gender, race, and age. Luccioni et al. ~\cite{luccioni2023stable} identify and quantify social biases in images generated by popular text-to-image models like DALL-E 2, and Stable Diffusion v1.4 and 2. Maluleke et al. ~\cite{maluleke2022studying} study racial biases in GANs and find that GANs mimic the racial distribution in the training data. 

\noindent \textbf{Debiasing generative models by retraining.} This line of work focuses on mitigating biases in generative models by retraining them \cite{nam2023breaking, xu2018fairgan, van2021decaf, sattigeri2019fairness, yu2020inclusive, choi2020fair, teo2023fair, um2023fair}. Some of these works \cite{xu2018fairgan, sattigeri2019fairness, van2021decaf} assume knowledge of the labels of the sensitive attribute and then debias the models such that there is no correlation between the decision attribute and the sensitive attribute. IMLE-GAN \cite{yu2020inclusive} ensures coverage of minority groups by combining GAN adversarial training with Implicit Maximum Likelihood Estimation (IMLE) \cite{li2018implicit}. Another body of works employs a balanced unlabelled reference dataset to ensure unbiased generations \cite{choi2020fair, teo2023fair, um2023fair}. Choi et al. \cite{choi2020fair} use a density-ratio based technique to identify the bias in datasets via the reference dataset, and learn a fair model based on importance reweighting with the help of both the original biased and the reference dataset. To capture the distance between the small reference data and the generated data, Um et al. \cite{um2023fair} use the LeCam Divergence \cite{le2012asymptotic}. On the other hand, Teo et al. \cite{teo2023fair} introduce a transfer learning approach to solve this problem by training the model on the biased dataset first and then adapting the model to the reference set.

\noindent \textbf{Debiasing generative models without training.}
As training of GANs and DMs can be resource-consuming, many methods prefer fair generation of images without explicit training \cite{ramaswamy2021fair, tan2020improving, melzi2023gandiffface, humayun2021magnet}. MaGNET~\cite{humayun2021magnet} aims to produce uniform sampling on the learned manifold of any generative model like GANs or VAEs, while Ramaswamy et al. ~\cite{ramaswamy2021fair} and Tan et al. ~\cite{tan2020improving} manipulate the latent space of GANs to generate balanced outputs. Controlling the generation with latent manipulation is easier in GANs due to its highly disentangled latent spaces in GANs~\cite{shen2020interpreting,parihar2022everything,hsr}. GANDiffFace~\cite{melzi2023gandiffface}, on the other hand, generates balanced synthetic for face recognition by first generating high-quality images from different demographics using GANs and then finetuning Stable Diffusion~\cite{ldm} using DreamBooth~\cite{ruiz2023dreambooth} to generate more images of such identities with different poses, expressions, etc. Multiple works attempt to mitigate biases in vision-language models and text-conditioned diffusion models as well \cite{chuang2023debiasing, zhang2023iti, seth2023dear, zhu2023debiased, berg-etal-2022-prompt}. Chuang et al. ~\cite{chuang2023debiasing} debias the text embedding using a calibrated projection matrix and shows unbiased generations without any additional training or data. However, debiasing the unconditional DMs has received less attention, which is the main focus of this work. 

\noindent \textbf{Guidance in Diffusion Models}. One of the primary techniques to condition the diffusion model is to guide the reverse diffusion model with the gradients of additional network ~\cite{clf_ddpm}. GLIDE~\cite{nichol2021glide} used CLIP~\cite{clip} based guidance for open world caption to image generation. Guidance is used for image-to-image translation~\cite{wolleb2022diffusion_img2img_med}, counterfactual generation~\cite{sanchez2022diffusion_counterfactual_gen}. However, guidance in its original form requires retraining of guiding networks on the noisy data from scratch. Few works overcome this by learning a mapping function from the diffusion feature space to sketches for sketch-guidance ~\cite{voynov2023sketch} and a universal guidance that repurposes pretrained networks for guidance ~\cite{bansal2023universal}.   
\vspace{-2mm}
\section{Method}
\vspace{-2mm}
We assume a setting where we are given a pretrained DM trained on \textit{biased} data and a reference distribution of the sensitive attributes $\mathbf{p^a_{ref}}$. Our goal is to generate data from the DM, whose generated attribute distribution $\mathbf{p^a_{\theta}}$ best approximates the reference $\mathbf{p^a_{ref}}$ without retraining. The key idea is to \textit{jointly} guide the denoising of a batch of samples such that $\mathbf{p^a_{ref}}$ $\approx$ $\mathbf{p^a_{\theta}}$. Directly computing $\mathbf{p^a_{\theta}}$ in a closed form is intractable. Instead, we train an \textbf{Attribute Distribution Predictor}, a linear projection that maps the intermediate batch features from \textit{h-space} of a denoising network to an estimate of attribute distribution $\mathbf{\hat{p}^a_{\theta}}$. 


\subsection{Preliminary}
\vspace{-2mm}
\textbf{Diffusion models} have emerged as a powerful family of generative models trained to learn the data distribution by gradual denoising from a Gaussian distribution. Starting from a clean point $\mathbf{x_0}$, and a set of scalar values $\{\alpha_t\}_{t=1}^T$, applying $\mathbf{t}$ steps of the forward diffusion process yields a noisy data point $\mathbf{x_t}$, where $\bar{\alpha_t}$ = $\prod_{i=1}^t \alpha_i$ and

\vspace{-3mm}
\begin{equation}
    \mathbf{x_t} = \sqrt{\bar{\alpha_t}}\mathbf{x_0} + (\sqrt{1-\bar{\alpha_t}})\epsilon, \epsilon \approx \mathcal{N}(0,I)
\vspace{-2mm}
\end{equation}
\vspace{-4mm}

\noindent 
A diffusion model is learned as a neural network $\epsilon_{\theta}$ that predicts the noise from given $\mathbf{x_t}$ and $\mathbf{t}$. The reverse process takes the form $q(\mathbf{x_{t-1}}|\mathbf{x_t}, \mathbf{x_0})$, which is parameterized as a Gaussian distribution. In this work, we consider DDIM \cite{ddim} sampling which first computes an estimate of the clean data point $\mathbf{\hat{x}_0}$ and then sample $\mathbf{x_{t-1}}$ from $q(\mathbf{x_{t-1}}|\mathbf{x_t}, \mathbf{\hat{x}_0})$. 

\noindent 
\textbf{Classifier guidance} is proposed to condition a diffusion model on class labels with the help of a pretrained classifier~\cite{clf_ddpm}. Specifically, a classifier $f_\phi(c|\mathbf{x_t},t)$ is trained on noisy images to predict the class label $c$. The gradients of the classifier are used to guide the diffusion sampling process to generate an image of the prescribed class $c$. Concretely, the classifier guidance performs sampling by updating the noise prediction 
$\epsilon_{\theta}(\mathbf{x_t}, \mathbf{t})$ as follows: 

\vspace{-6mm}
\begin{equation}
    \hat{\epsilon}_{\theta}(\mathbf{x_t}, t) = \epsilon_{\theta}(\mathbf{x_t}, t) - \sqrt{1-\alpha_t}\nabla_{\mathbf{x_t}}\log f_{\phi}(c|\mathbf{x_t},t)
\label{eq:eq2}
\end{equation}
\vspace{-8mm} 


\subsection{Classifier guidance for debiasing}
\label{subsec:clf_guide}
\begin{figure}[h]
    \centering
    \includegraphics[width=1\linewidth]{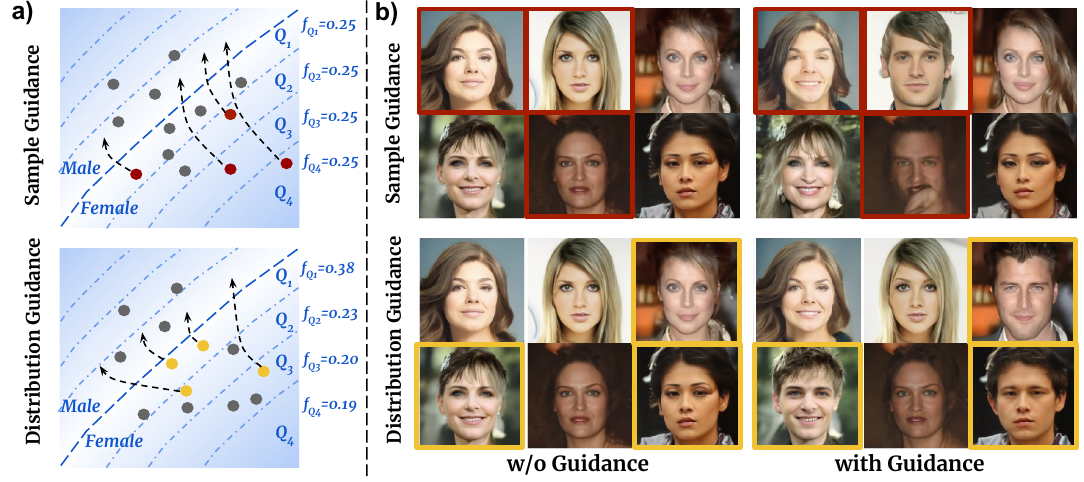}
    \vspace{-7mm}
    \caption{\textbf{Sample guidance vs Distribution guidance}. \textbf{a)} After a few steps of denoising ($t=\tau$), the generated samples have learned some discriminative features for gender. Sample guidance randomly selects \textcolor{BrickRed}{samples} from the batch uniformly from all the quantiles for conversion enforcing samples with dominant female features to also change. However, distribution guidance majorly converts the \textcolor{Goldenrod}{samples} close to the decision boundary (Q1/Q2), which is \textit{easy} to convert. 
    \textbf{b)} Visualization of the generated samples. Samples transformed with distribution guidance are natural looking without any distortion, whereas images with sample guidance suffer from distortion or unnatural appearance.}
    \vspace{-3mm}
    \label{fig:sample_distribution_compare} 
\end{figure}

\vspace{-2mm}
A promising approach is to leverage pretrained attribute classifiers to guide towards balanced generation. Assuming a  reference attribute distribution $\mathbf{p^{a}_{ref}}$ for a binary attribute $\mathbf{a}$ (e.g., gender) and corresponding attribute classifier ${f}_{\phi}$, we parameterize $\mathbf{p^{a}_{ref}}$ as a Bernoulli distribution with parameter $\mathbf{r}$ denoting fraction of \textit{males} samples. To generate samples following $\mathbf{p^{a}_{ref}}$, we can randomly select $\mathbf{Nr}$ samples from the batch size of $\mathbf{N}$ and guide them towards \textit{male} class using predictions from ${f}_{\phi}$. Similarly, for the remaining $\mathbf{N(1-r)}$ samples, we guide them towards \textit{female} class. In practice, such a \textit{sample guidance} follows $\mathbf{p^{a}_{ref}}$ up to some extent, but results in inferior sample quality (see Fig. {\ref{fig:sample_distribution_compare}}b). 

\vspace{1mm}
\noindent
\textbf{\textit{\textcolor{blue}{Insight.1}}}: {\textit{Transforming samples close to the decision boundary is easier and results in higher quality generation.}}

\begin{figure}[h]
    \centering
    \includegraphics[width=1\linewidth]{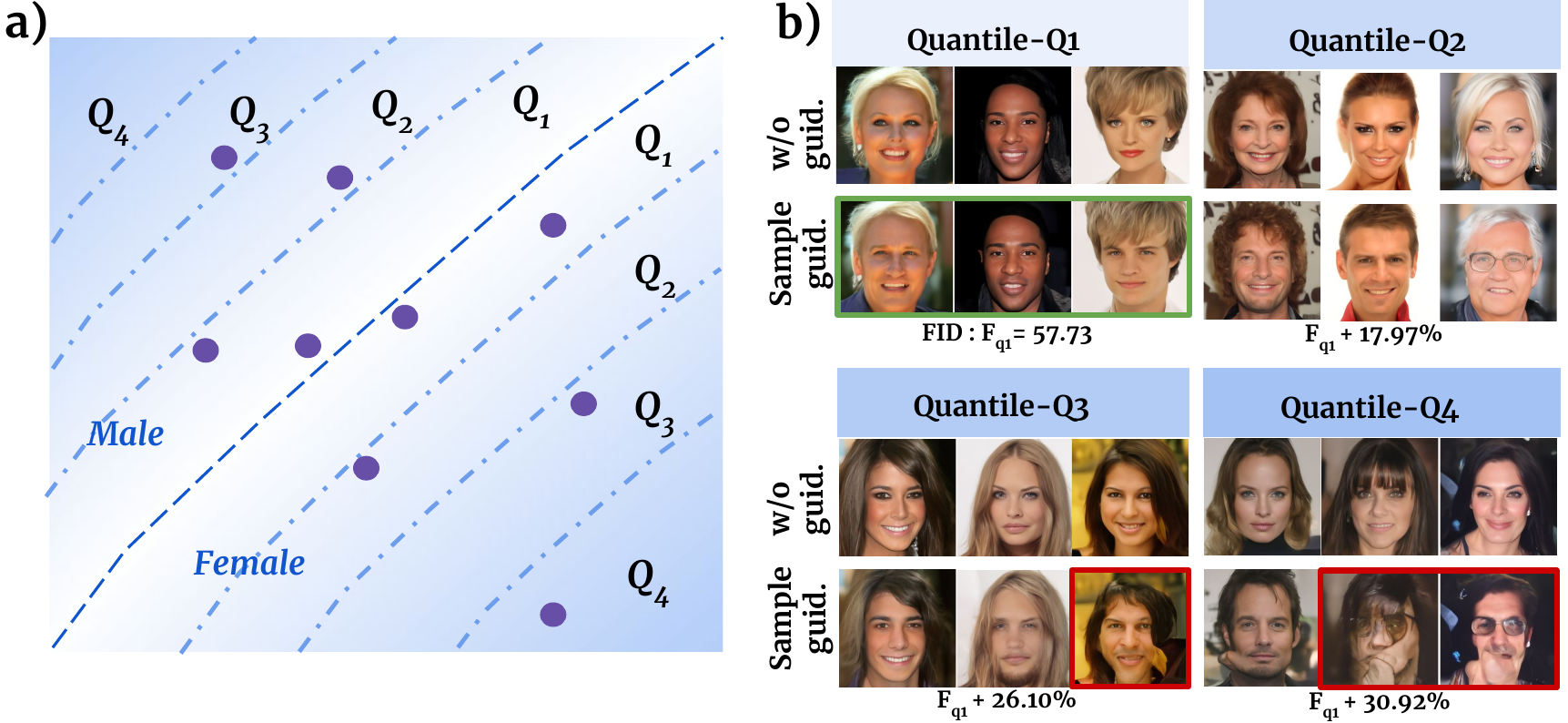}
    \vspace{-7mm}
    \caption{Distribution of samples w.r.t. decision boundary}
    \vspace{-5mm}
    \label{fig:quantile-plot} 
\end{figure}

\vspace{1mm}
\noindent
\textbf{Remark:} We performed an insightful experiment for changing gender attribute (\textit{female} to \textit{male}) using classifier guidance. We group the samples in four quantiles based on their distance from the decision boundary of a pre-trained gender classifier in Fig.~\ref{fig:quantile-plot}a). Next, we perform sample-based guidance over the samples from each quantile in Fig.~\ref{fig:quantile-plot}b). The samples which are close to the decision boundary (quantiles Q1 and Q2) are easily transformed with guidance, whereas the samples away from the decision boundary (quantiles Q3 and Q4) are distorted during the guidance process. This is also quantified with the quantile-wise FID against a real set (of \textit{male} images), where the samples from Q1/Q2 have better FID after conversion.

\vspace{-3mm}

\vspace{1mm}
\noindent
\textbf{\textit{\textcolor{blue}{Insight.2}}}: {\textit{Attempting to steer the generation of individual samples towards a pre-defined attribute class is overly restrictive and leads to inferior generation quality.
}}

\vspace{1mm}
\noindent 
\textbf{Remark:} Sample guidance requires enforcing a preset attribute state (\textit{male}/\textit{female}) to each sample of a batch during the denoising process, which is too stringent and results in distorting the outputs. This is particularly bad in the case when samples from quantiles Q4 are selected for transformation, as shown in Fig.~\ref{fig:quantile-plot}. Given an intermediate time-stamp $\tau$, the samples in the earlier stages ($t>\tau$) of \textit{reverse} diffusion process are close to the noise space, resulting in the poor classifier of the corresponding stages. The guidance becomes effective only in the later stage of denoising ($t<\tau$). However, till timestep $\tau$, some facial features are already formed~\cite{meng2022sdedit}. Hence, enforcing a preset attribute state at later stages ($t<\tau$) is restrictive and results in \textit{sample collapse} (Fig.~\ref{fig:sample_distribution_compare}-b)). For e.g., for gender attribute, if a sample has formed dominant \textit{female} features till $\tau$ it will be in Q4, and enforcing it to be $male$ is overly restrictive, resulting in distorted generations.  


\vspace{-3mm}

\subsection{Distribution Guidance}
\vspace{-2mm}
We propose an alternate guidance strategy termed \textit{Distribution Guidance} for \textit{fair} generation, which provides more flexibility to modify the attribute states during the denoising process. The key idea is to \textit{jointly} denoise a batch of samples $\mathbf{x_T^{[1:N]}}$, with the guidance from reference attribute distribution $\mathbf{p^a_{ref}}$. To realize distribution guidance, we define a differentiable distribution prediction function $g_\psi$ that maps the batch samples to an estimate of generated attribute distribution $\mathbf{\hat{p}^a_{\theta}}$. We learn the function $g_\psi$ over an intermediate feature space - \textit{h-space}~\cite{h_space} of denoising network instead of image space for efficiency (Sec.\ref{subsec:guide_h_space}). Hence, the batch estimate is given by, $\mathbf{\hat{p}^a_\theta} = g_\psi(\mathbf{h_t^{[1:N]}},\mathbf{t})$, where $\mathbf{h_t^{[1:N]}}$ is the bottleneck U-Net feature representation of the batch samples $\mathbf{x_t^{[1:N]}}$.
Further, we define a loss function $\mathcal{L}(.)$ that measures the similarity of two distributions. During denoising we guide the batch of samples to bring $\mathbf{\hat{p}^a_{\theta}}$ closer to the reference, i.e., $\mathcal{L}(\mathbf{\hat{p}^a_{\theta}}, \mathbf{p^A_{ref}}) \approx 0$. This can be easily integrated into the reverse diffusion process as an extension of classifier guidance by modifying Eq.\ref{eq:eq2}.    


\vspace{1mm}
\noindent \textbf{\textit{\textcolor{blue}{Insight.3}}}: \textit{Distribution guidance provides flexibility to batch samples and transforms easier samples close to the decision boundary to match the required distribution.}
\vspace{1mm}  


\noindent \textbf{Remark:} As distribution guidance does not require presetting for attribute states \textit{for each sample}, it gives more flexibility during the generation as long as $\mathbf{\hat{p}^a_{\theta}}$ follows $\mathbf{p^a_{ref}}$. \Eg, in our running example of gender attribute, only those samples will change gender after $t=\tau$, which are close to the decision boundary (in Q1 \& Q2). In contrast, in sample guidance, random samples are forced to change the attribute state (Fig.~\ref{fig:sample_distribution_compare}) and are spread equally in all four quantiles, resulting in inferior quality. This is quantified by the fraction of samples being transformed ($f_{Qi}$) with each guidance in Fig.~\ref{fig:sample_distribution_compare}a), where distribution guidance majorly transforms samples from Q1 \& Q2 and results in \textit{fair} and high-quality samples (Fig.~\ref{fig:sample_distribution_compare}b)).

A major design decision for implementing distribution guidance is the selection of function $g_\psi$. The conventional approach is to train multiple attribute classifiers at each level of noise in the image space. Instead, we propose to leverage intermediate semantic features from the diffusion model itself and repurpose them for the task of attribute classification. This framework is very efficient as compared to image space classifiers.

\subsection{Guidance in the H-space}
\label{subsec:guide_h_space}
\begin{figure}[h]
    \centering
    \includegraphics[width=0.8\linewidth]
    {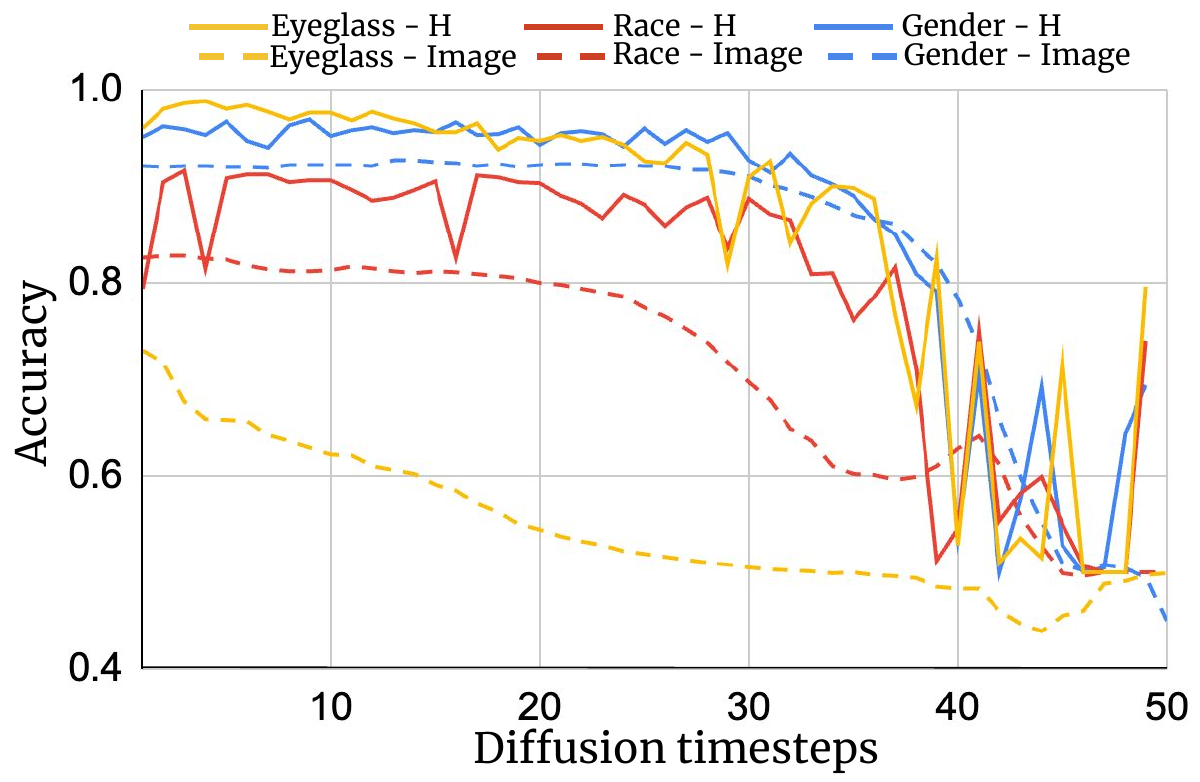}
    \vspace{-4mm}
    \caption{Classification accuracy for linear \textit{h-space} classifiers and ResNet-18 image space classifiers trained on $2K$ training examples. \textit{h-space} classifiers are data efficient and achieve superior performance even with a linear layer}
    \vspace{-2mm}
    \label{fig:h-space-clf-evals} 
\end{figure}

\vspace{-2mm}
Diffusion models, although trained for image generation, learn semantically rich representations in the internal features of denoising network $\mathbf{{\epsilon_\theta}}$. These representations have been successfully used for segmentation~\cite{diff_segm} and classification~\cite{h_space_clf}. Motivated by this, we ask \textit{can we use these internal features for training guidance classifiers?} To answer this, we take features from the bottleneck layer of the diffusion U-Net, termed as the \textit{h-space}~\cite{h_space} and attach a linear head for classification. The trained classifiers achieve good classification performance across multiple attributes, as shown in Fig.~\ref{fig:h-space-clf-evals}. Hence, we use the trained $h-space$ classifiers to realize $g_\psi$. Specifically, we train a network Attribute Distribution Predictor that maps the batch \textit{h-space} features to an estimate of attribute distribution $\mathbf{\hat{p}^a_{\theta}}$.

\vspace{-3mm}
\begin{figure}[h]
    \centering
    \includegraphics[width=1\linewidth]{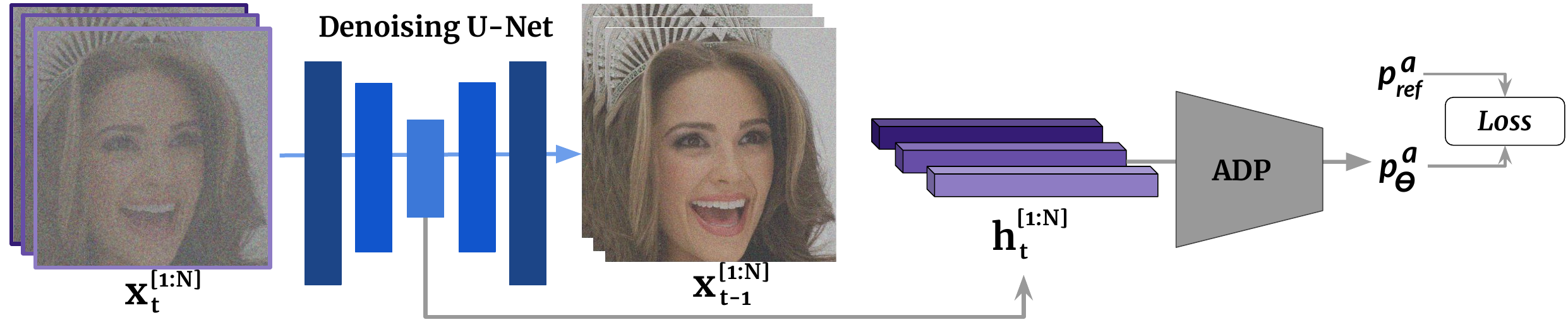}
    \vspace{-7mm}
    \caption{\textbf{Distribution guidance in the \textit{h-space}}. For a given batch $\mathbf{x_t^{[1:N]}}$, we extract the intermediate \textit{h-space} representation $\mathbf{h_t^{[1:N]}}$ and pass it through ADP to obtain attribute distribution $\mathbf{\hat{p^a_\theta}}$. Guidance updates $\mathbf{h_t^{[1:N]}}$ by backpropagating the derivative of loss.}
    \vspace{-3mm}
    \label{fig:h-space-clf} 
\end{figure}

\noindent \textbf{Attribute Distribution Predictor (ADP)} is realized via a linear attribute classifier conditioned on the diffusion time step $\mathbf{t}$. Given a batch of generating samples $\mathbf{x_t^{[1:N]}}$, we extract the corresponding \textit{h-space} features $\mathbf{h_t^{[1:N]}}$ = $\epsilon^E_\theta(\mathbf{x_t^{[1:N]}}, \mathbf{t})$, where $\epsilon^E_\theta$ is the encoder of U-Net network. Next, we pass the batch $\mathbf{h_t^{[1:N]}}$ to the attribute classifier and obtain a batch of softmax predictions. The softmax predictions are aggregated per class to obtain the attribute distribution estimate $\mathbf{\hat{p}^a_\theta}$. Finally, we update the intermediate \textit{h-vectors} with the gradients of the distribution matching loss $\mathcal{L}$, where $\mathbf{\gamma}$ is the guidance strength parameter:

\vspace{-4mm}
    \begin{align*}
    \mathbf{\tilde{h}_t^{[1:N]}} = \mathbf{h_t^{[1:N]}} & - \gamma * \nabla_{\mathbf{h_t^{[1:N]}}} \mathcal{L}(\mathbf{\hat{p}^a_\theta}, \mathbf{p^a_{ref}}) \\ 
     \mathbf{\hat{p}^a_\theta} = & g_\psi(\mathbf{h_t^{[1:N]}}, \mathbf{t}) 
    \end{align*} 

\noindent Finally, we obtain the distribution guided noise predictions $\tilde{\epsilon}$ by passing the $\mathbf{h^{[1:N]}_t}$ through the U-Net decoder $\epsilon^D_\theta$, \ie, $\tilde{\epsilon}(\mathbf{x^{[1:N]}_t}, \mathbf{t})=\epsilon^D_\theta(\mathbf{h^{[1:N]}_t}, \mathbf{t})$. The predicted noise is then used to update batch $x^{[1:N]}_{t-1}$ using DDIM~\cite{ddim}.


\vspace{2mm} 
\noindent \textbf{\textit{\textcolor{blue}{Insight.4}}}:
\textit{H-space guidance is extremely effective and efficient as compared to image space guidance.}
\vspace{1mm}

\noindent \textbf{Remark.} As compared to conventional image space guidance, guidance in h-space has multiple advantages: \textbf{i)} it requires only a set of linear layers to implement the classifiers, \textbf{ii)} it is fast to backpropagate during guidance as compared to image models, \textbf{iii)} highly data-efficient and can be trained with only a few thousand examples due to semantically rich \textit{h-space}. In the experiments section, we compare these properties of the \textit{h-space} guidance.

\vspace{-3mm}

\section{Experiments}
\label{sec:exp}
\vspace{-2mm}
In this section, we first discuss the evaluation metrics, followed by the implementation details. We take face datasets as the subject of study, as they are subject to very high demographic biases. We compare our method with other inference time debiasing approaches for both single-attribute and multi-attribute debiasing. Further, we present detailed ablations to validate design choices. Finally, we show the generalization of our debiasing approach on large text-to-image generation model Stable Diffusion~\cite{ldm}. 

\subsection{Evaluation Metrics}
\label{subsec:metrics} 
\vspace{-2mm}
A \textit{fair} generative model is evaluated on two grounds: image quality and fairness. We discuss the metrics used to measure these two aspects of the generated images below.

\noindent \textbf{Fairness}. The primary goal of this paper is to generate images based on the reference distribution $\mathbf{p_{ref}^a}$. We follow Choi et al. ~\cite{choi2020fair} to define the \textbf{Fairness Discrepancy} (FD) metric. Given an attribute $\mathbf{a}$, we assume access to a high-accuracy classifier for $\mathbf{a}$ (denoted as $\mathcal{C}_a$), and using the predictions from the latter, we compute the following~\cite{choi2020fair, teo2023fair}:

\vspace{-3mm}
$$||\bar{p} - \mathbb{E}_{\mathbf{x} \sim p_\theta(\mathbf{x})}(\mathbf{y})||_2$$
\vspace{-5mm}

\noindent 
where $\mathbf{y}$ is the softmax output of the classifier $\mathcal{C}_a(\mathbf{x})$, $\bar{p}$ is a uniform vector of the same dimension as $\mathbf{y}$, $p_\theta$ is the distribution of the generated images. The lower the FD score, the closer the distribution of the attribute values is to the uniform distribution -- i.e., the generated images are fairer with respect to attribute $\mathbf{a}$.

\noindent \textbf{Image Quality}. To measure the quality of generation, we follow the standard Fréchet Inception Distance (FID)~\cite{heusel2017gans}. We compute this metric with the help of an attribute-wise balanced dataset sampled from the original training data. 

\vspace{-1mm}
\subsection{Implementation Details}
\label{subsec:implementation}
\vspace{-2mm}
\textbf{Training h-space classifiers.} We start with creating a paired dataset $\mathcal{D}^h_{clf}$ of h-vector and attribute labels. Specifically, we take a subset $\mathcal{D}$ of the CelebA-HQ~\cite{pro-gan} dataset and obtain attribute labels for the images using the pretrained attribute classifier $\mathcal{C}_a$. Next, we embed image $\mathbf{I^i} \in \mathbf{D}$ to obtain the corresponding h-space representation $\mathbf{\mathcal{H}^i} = \{\mathbf{h^i_t}\}^{t=T}_{t=0}$ using DDIM~\cite{ddim} inversion. This yields a labelled dataset $\mathcal{D}^h_{clf}$ with pairs $(\mathcal{H}^i, \mathbf{y^i})$, where $\mathbf{y^i} = \mathcal{C}_a(I^i)$ is the predicted attribute label for image $\mathbf{I^i}$ (\eg \textit{male / female}). Next, we train \textit{h-space} attribute classifiers $\mathcal{C}^h_a(\mathbf{h_t}, t)$ as a \textit{linear} head over $\mathbf{h_t}$ and conditioned on time $t$ (ignored sample index $i$ for brevity). These obtained classifiers generate high-accuracy attribute predictions, as shown in Fig.~\ref{fig:h-space-clf-evals}. Further details about the dataset and classifiers are provided in Sec.\textcolor{red}{D} of supplement. 

\begin{figure*}
    \centering
    \includegraphics[width=\linewidth]{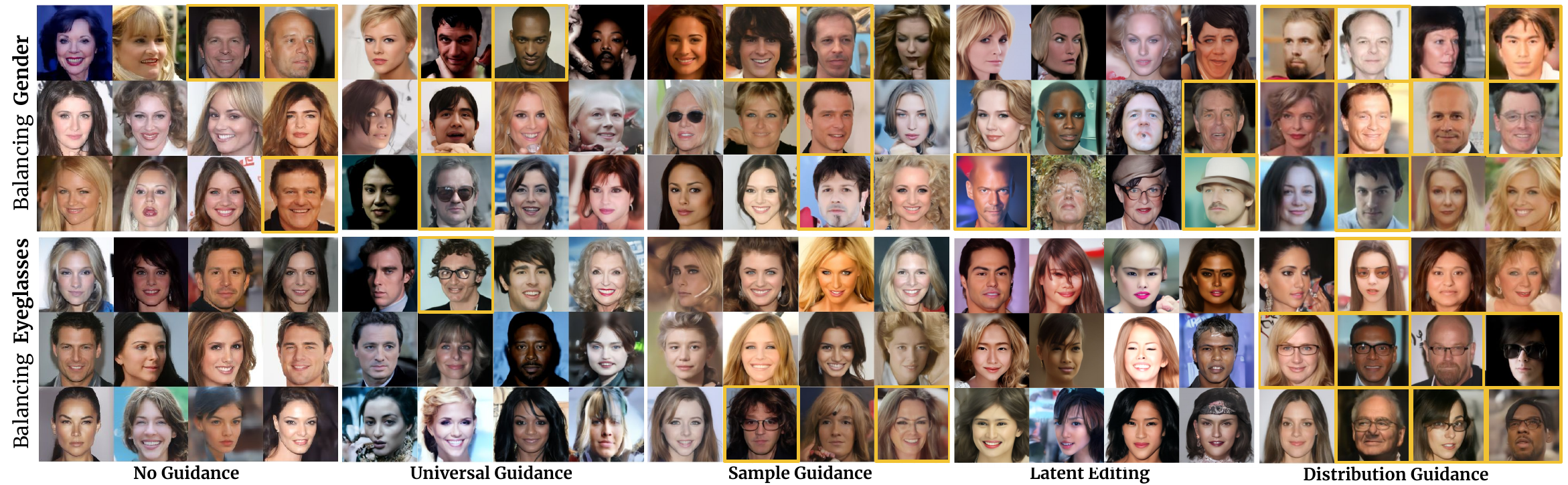} 
    \vspace{-6mm}
    \caption{\textbf{Balancing generated data across \textit{gender} and \textit{eyeglass} attributes with various guidance strategies.} Minority group \textcolor{Goldenrod}{$male$} and \textcolor{Goldenrod}{$eyeglasses$} are marked. Image-space \textbf{Universal Guidance} and \textbf{Sample Guidance} results generate imbalanced and poor-quality images. For \textit{gender}, Sample Guidance is able to generate only a few \textit{males} while maintaining quality. \textbf{Latent Editing} produces collapsed images for gender and fails to generate eyeglasses. The proposed \textbf{Distribution Guidance} balances (close to the ratio of $0.50:0.50$) both the attributes and generates high-quality images.}
    \vspace{-6mm} 
    \label{fig:single-debias}
\end{figure*}

\noindent 
\textbf{Distribution guidance.} To perform distribution guidance, we realize the distribution prediction function, \textbf{ADP} with \textit{h-space} classifiers $\mathcal{C}^h_a$. Specifically for a generating batch $\mathbf{x_t^{[1:N]}}$, we obtain the corresponding $h-space$ representation $\mathbf{h_t^{[1:N]}}$ and obtain a set of attribute predictions $\mathbf{\hat{y}^{[1:N]}_t}$ from classifiers $\mathcal{C}^h_a$. Finally, we add all the softmax values for all the $N$ images in the batch for each class to obtain the estimate of $\mathbf{p^a_{\theta}}$. We use \textit{Chi-square} distance as the loss $\mathcal{L}$ with the reference distribution $\mathbf{p^a_{ref}}$. 

\noindent \textbf{Diffusion Model Architecture.} We evaluated our proposed debiasing method on two state-of-the-art pretrained DMs: a) an unconditional DM, P2~\cite{choi2022perception}, trained on CelebA-HQ dataset, and b) a text conditional DM StableDiffusion v1.5~\cite{ldm} trained on the LAION~\cite{schuhmann2022laion} dataset. Both these models have exceptional image generation quality; however, they have significant bias concerning the sensitive face attributes, as shown in the following subsections. 

\vspace{-1mm} 
\subsection{Baselines}
\vspace{-2mm}
We compare our proposed method against two techniques for guidance-based generations for DMs, one Latent-based editing method~\cite{h_space}, and a state-of-the-art sampling-based technique for debiasing generative models, MagNet~\cite{humayun2021magnet}.

\noindent \textbf{Sample guidance.} We use classifier guidance as explained in Sec.~\ref{subsec:clf_guide} in the \textit{h-space} using the trained \textit{h-space} attribute classifiers $\mathcal{C}_a^h$. Such guidance requires presetting an attribute state for each generating image and pushing the trajectory to change the attribute state. 

\noindent \textbf{Universal guidance}~\cite{bansal2023universal} performs guidance in the image space but uses pretrained classifiers trained on the clean image. This resolves the additional requirement of training image classifiers on noisy images. The key idea is to use the DDIM scheduler and predict the approximation of $\mathbf{\hat{x_0}}$ from noise image $\mathbf{x_t}$, and pass it through pretrained image classifier $\mathcal{C}_A$. However, this process has two shortcomings: it is slow as it backpropagates the gradients from the image classifier (Experimentally, we found it to be  $7$ times slower than \textit{h-space} guidance) and it performs poorly at the early stage due to an inaccurate approximation of $\mathbf{x_0}$. We use two settings where we vary the number of images in the training set of the image space attribute classifier: (1) Using the same training set $|\mathcal{D}_{clf}| = 2K$ as the \textit{h-space} classifier and (2) Using the entire CelebA-HQ dataset, i.e. 30k images.

\noindent \textbf{Latent-based Editing}~\cite{h_space} generates images with a specific set of attributes. 
Such a technique is popularly used in debiasing GANs~\cite{ramaswamy2021fair, karakas2022fairstyle} because of the well-known disentangled latent spaces of GANs that allow for such edits. Recent works have shown ~\cite{h_space} that similar semantic control is also present in the \textit{h-space} of DMs and can be used for latent-based editing. We capitalize on this finding and perform latent editing in the \textit{h-space} to generate images of desired attributes for \textit{fair} generation. 

\noindent \textbf{MagNet}~\cite{humayun2021magnet} is an unsupervised method enabling \textit{fair} sampling from a pretrained model. They propose a method for uniform sampling on the image manifold to generate under-represented groups equally. We generated results by MagNet sampling from a StyleGAN2~\cite{sg2} model trained on the FFHQ~\cite{ffhq} dataset from their official codebase. Notably, as the base model for MagNet is StyleGAN2, we cannot directly compare FIDs with our DM debiased results and report random generations from StyleGAN2 as a reference. 

\subsection{Main Results}
\vspace{-2mm}

We first present the quantitative and qualitative results on debiasing single binary attributes. Second, we debias multiple attributes simultaneously. We finally present the case of debiasing attributes with multi-class labels.

\begin{table}[]
\caption{Evaluation of \textit{balanced} generation for single attribute}
\vspace{-3mm}
\begin{adjustbox}{width=0.45\textwidth,center}
\begin{tabular}{@{}c|cc|cc|cc@{}}
\toprule
                            & \multicolumn{2}{c|}{Gender}                                                      & \multicolumn{2}{c|}{Race}                                                        & \multicolumn{2}{c}{Eyeglasses}                                                   \\ \midrule
\multicolumn{1}{c|}{Method} & \multicolumn{1}{c|}{FD $\downarrow$}    & \begin{tabular}[c]{@{}c@{}}FID $\downarrow$\\ \end{tabular} & \multicolumn{1}{c|}{FD $\downarrow$}    & \begin{tabular}[c]{@{}c@{}}FID $\downarrow$\\ \end{tabular} & \multicolumn{1}{c|}{FD $\downarrow$}    & \begin{tabular}[c]{@{}c@{}}FID $\downarrow$\\ \end{tabular} \\ \midrule
Random Sampling             & \multicolumn{1}{c|}{0.178} & 54.59                                               & \multicolumn{1}{c|}{0.334} & 60.01                                               & \multicolumn{1}{c|}{0.251} & 75.21                                               \\
Universal Guidance (2k)~\cite{bansal2023universal}         & \multicolumn{1}{c|}{0.193} & 52.10                                               & \multicolumn{1}{c|}{0.377} & 93.42                                               & \multicolumn{1}{c|}{0.189} & 64.55                                               \\
Universal Guidance (30k)~\cite{bansal2023universal}         & \multicolumn{1}{c|}{0.127} &  \underline{48.94}                                               & \multicolumn{1}{c|}{0.326} &  58.52                                               & \multicolumn{1}{c|}{\underline{0.051}} & 78.57                                               \\
Latent Editing~\cite{h_space}           & \multicolumn{1}{c|}{\textbf{0.001}} & \multicolumn{1}{c|}{\textbf{37.40}}                         & \multicolumn{1}{c|}{0.214} & \multicolumn{1}{c|}{\textbf{42.69}}                         & \multicolumn{1}{c|}{0.330} & 75.04              \\
H-Sample Guidance (ours)       & \multicolumn{1}{c|}{0.113} & 51.46                                               & \multicolumn{1}{c|}{\underline{0.184}} & 56.53                                               & \multicolumn{1}{c|}{0.118} & \underline{57.63}                                               \\
H-Distribution Guidance (ours)     & \multicolumn{1}{c|}{\underline{0.049}} & 50.27                                            & \multicolumn{1}{c|}{\textbf{0.113}} & \underline{52.38}                                               & \multicolumn{1}{c|}{\textbf{0.014}} & \textbf{51.78}                                               \\ \midrule
StyleGAN2 - Random sampling            & \multicolumn{1}{l|}{0.307} & \multicolumn{1}{l|}{112.28}                         & \multicolumn{1}{c|}{0.463} & \multicolumn{1}{c|}{123.97}                         & \multicolumn{1}{c|}{0.276} & \multicolumn{1}{c}{117.83}                          \\
StyleGAN2 - Magnet~\cite{humayun2021magnet}                    & \multicolumn{1}{l|}{0.267} & \multicolumn{1}{l|}{91.15}                          & \multicolumn{1}{c|}{0.454} & \multicolumn{1}{c|}{97.05}                          & \multicolumn{1}{c|}{0.281} & \multicolumn{1}{c}{106.55}                          \\ \bottomrule
\end{tabular}
\end{adjustbox}
\label{tab:single_attr}
\end{table}

\noindent \textbf{Quantitative evaluation.} We evaluate our debiasing method for the single attribute case by generating balanced generations of individual sensitive attributes - \textit{gender}, \textit{eyeglasses}, and \textit{race}, in Tab.~\ref{tab:single_attr}. As these attributes are binary, the synthesized images are expected to have a $1:1$ ratio of the sensitive attributes (for e.g. $0.50$ fraction of males and $0.50$ fraction of females in case of gender). Specifically, we generate $10K$ images from each method per attribute and compute the metrics defined in Sec.~\ref{subsec:metrics}. For most attributes, the proposed guidance method outperforms all the baselines in terms of visual quality as measured by FID and bias metric measured by FD. Although Latent editing has a better FID and FD for gender, on qualitative evaluation (as elaborated in the next section), artifacts are seen in the images (Fig.~\ref{fig:single-debias}). Moreover, this methods fails to mitigate bias in case of multiple attributes (Tab.~\ref{tab:multi_attr}). Sample guidance achieves comparable FID for \textit{gender}; however, higher FD indicates inferior debiasing. The tradeoff between FD and FID is discussed in Sec.\textcolor{red}{A.1} of supplement. This supports our thesis that the distribution guidance provides enough flexibility during generation, resulting in high-quality outputs even with a high guidance scale.

\noindent
\textbf{Qualitative evaluation.} We present results for balancing \textit{gender} and \textit{eyeglasses} attributes in Fig.~\ref{fig:single-debias}. We randomly sample $20$ starting noise and use individual guidance methods for debiasing. Without guidance, the DM mostly generates \textit{female} faces for the gender attribute. Although latent editing achieved better quantitative metrics, it produces images with artifacts and leads to collapsed results. Although Sample and Universal guidance increase the number of \textit{males}, some images collapse. On the other hand, distribution guidance generates an almost equal number of males and females without affecting the generation quality. Moreover, all the baselines are not able to generate \textit{eyeglasses}, whereas our method leads to highly balanced generations.

\noindent \textbf{Multiple attributes.} We apply our method for debiasing multiple attributes simultaneously in Tab.~\ref{tab:multi_attr}. Specifically, given two reference distributions  $\mathbf{p^{a_1}_{ref}}$ and $\mathbf{p^{a_2}_{ref}}$, we add guidance from two pretrained attribute distribution predictors $g_{\psi_1}$ and $g_{\psi_2}$. For this experiment, we define both the reference distribution as uniform for each attribute ($50\%$-$50\%$ splits). The generated results follow the reference, resulting in a balanced generation across attributes. Further analysis is provided in Sec.\textcolor{red}{B} of supplement.

\begin{table}[]
\vspace{-4mm} 
\caption{Evaluation of \textit{balanced} generation for multiple attribute}
\vspace{-3mm}
\begin{adjustbox}{width=0.45\textwidth,center}
\begin{tabular}{@{}c|cc|cc|cc@{}}
\toprule
                            & \multicolumn{2}{c|}{Gender + Race}                                               & \multicolumn{2}{c|}{Eyeglasses + Race}                                           & \multicolumn{2}{c}{Gender + Eyeglasses}                                          \\ \midrule
\multicolumn{1}{c|}{Method} & \multicolumn{1}{c|}{FD $\downarrow$}    & \begin{tabular}[c]{@{}c@{}}FID $\downarrow$\\ \end{tabular} & \multicolumn{1}{c|}{FD $\downarrow$}    & \begin{tabular}[c]{@{}c@{}}FID $\downarrow$\\ \end{tabular} & \multicolumn{1}{c|}{FD $\downarrow$}    & \begin{tabular}[c]{@{}c@{}}FID $\downarrow$\\ \end{tabular} \\ \midrule
Random Sampling             & \multicolumn{1}{c|}{0.256} & 60.68                                               & \multicolumn{1}{c|}{0.292} & 89.14                                               & \multicolumn{1}{c|}{0.214} & 70.97                                               \\
Latent Editing~\cite{h_space}         & \multicolumn{1}{c|}{\underline{0.124}} & 64.84                                               & \multicolumn{1}{c|}{0.219} & 90.63                                               & \multicolumn{1}{c|}{0.230} &  74.93                                               \\
Universal Guidance (2k)~\cite{bansal2023universal}        & \multicolumn{1}{c|}{0.283} & 71.84                                               & \multicolumn{1}{c|}{0.264} & 91.54                                               & \multicolumn{1}{c|}{0.157} & 80.57                                               \\
H-Sample Guidance (ours)       & \multicolumn{1}{c|}{0.241} & \underline{59.78}                                               & \multicolumn{1}{c|}{\underline{0.135}} & \underline{67.87}                                               & \multicolumn{1}{c|}{\underline{0.079}} & \underline{52.03}                                               \\
H-Distribution Guidance (ours)   & \multicolumn{1}{c|}{\textbf{0.075}} & \textbf{49.91}                                               & \multicolumn{1}{c|}{\textbf{0.101}} & \textbf{57.46}                                               & \multicolumn{1}{c|}{\textbf{0.057}} & \textbf{47.45}                                               \\ \bottomrule
\end{tabular}
\end{adjustbox}
\label{tab:multi_class}
\end{table}

\noindent \textbf{Multi Class attributes.} We evaluate the efficacy of our approach in balancing multi-class attributes - \textit{age} and \textit{race} in Tab.~\ref{tab:multi_attr}. We use FFHQ~\cite{ffhq} dataset for both attributes and obtained annotations using pretrained models as ground truth labels are unavailable. For \textit{age} attribute, we use a pretrained VIT age classifier~\cite{vit-age-model} to produce $3$ classes: Young (${<20}$ yrs), Adult (${20-60}$ yrs) and Old (${>60}$ yrs). For \textit{race}, we use the Fairface race classifier~\cite{fairface-model} to obtain $4$ classes: White, Black, Asian, and Indian. Our method successfully debiases multi-class attributes and beats the random and sample guidance in all cases.

\subsection{Generating Imbalanced Distributions} 
\vspace{-2mm}
We test our distribution guidance for generating imbalanced attribute distribution by providing skewed $\mathbf{p^a_{ref}}$ - \textbf{i)} $0.20$ \textit{female} and $0.80$ \textit{male} \textbf{ii)} $0.10$ \textit{white race} and $0.90$ \textit{black race}. These two settings are extremely challenging given that \textit{male} and \textit{black race} are minority groups in the 

\begin{table}[h]
\caption{\textit{Balanced} generation for multi-class attribute}
\vspace{-3mm}
\begin{adjustbox}{width=0.36\textwidth,center}
\begin{tabular}{@{}c|cc|cc@{}}
\toprule
                            & \multicolumn{2}{c|}{Age (3 classes)}                                               & \multicolumn{2}{c}{Race (4 classes)}                                          \\ \midrule
\multicolumn{1}{c|}{Method} & \multicolumn{1}{c|}{FD $\downarrow$}    & \begin{tabular}[c]{@{}c@{}}FID $\downarrow$\\ \end{tabular} & \multicolumn{1}{c|}{FD $\downarrow$}    & \begin{tabular}[c]{@{}c@{}}FID $\downarrow$\\ \end{tabular} \\
\midrule
Random Sampling             & \multicolumn{1}{c|}{0.256} & 60.68                                               & \multicolumn{1}{c|}{0.292} & 89.14                                                                                       \\
H-Sample Guidance          & \multicolumn{1}{c|}{0.124} & 64.84                                               & \multicolumn{1}{c|}{0.219} & 90.63                                                                                            \\
H-Distribution Guidance        & \multicolumn{1}{c|}{0.283} & 71.84                                               & \multicolumn{1}{c|}{0.264} & 91.54                                                                                       \\ \bottomrule
\end{tabular}
\end{adjustbox}
\label{tab:multi_attr}
\end{table}

\begin{figure}[h]
    \centering
    \vspace{-2mm} 
    \includegraphics[width=\linewidth]{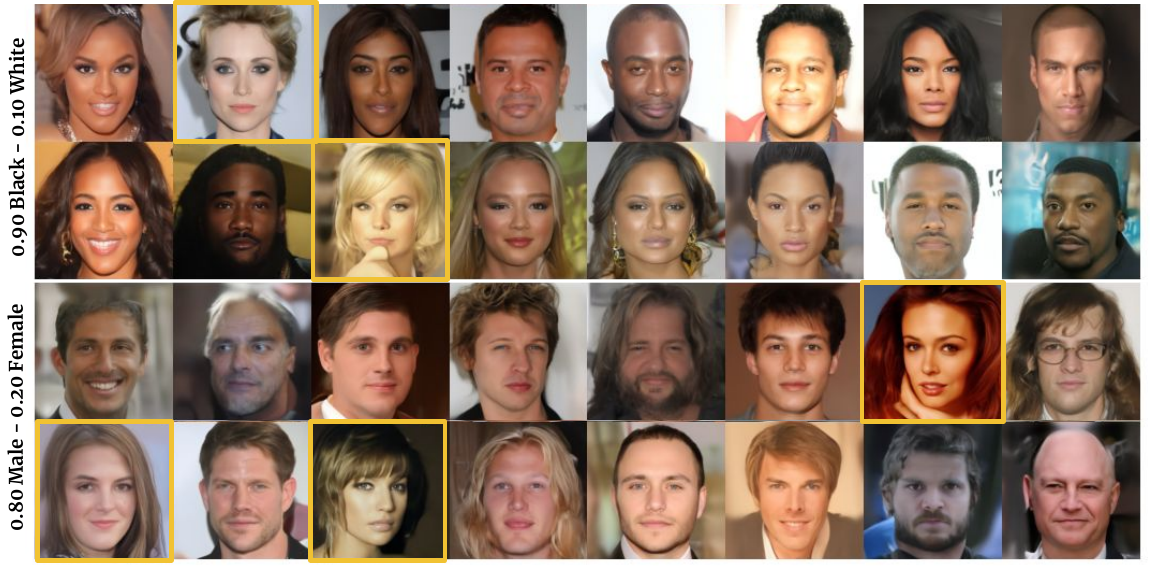}
    \vspace{-8mm}
    \caption{Generating imbalanced data with two non-uniform reference distribution - \textbf{i) 0.9B-0.1W}, \textbf{ii) 0.8M-0.2F.} \textcolor{Goldenrod} {Minority classes} are marked. Distribution guidance generates high-quality images and \textit{closely follows} the skewed reference distribution.}
    \vspace{-6mm}
    \label{fig:skew_generation} 
\end{figure}

\begin{table}[h]
\caption{Distribution guidance for imbalanced generation}
\vspace{-3mm}
\begin{adjustbox}{width=0.36\textwidth,center}
\begin{tabular}{@{}ccccc@{}}
\toprule
\multicolumn{1}{c}{}                         & \multicolumn{2}{l}{0.20F - 0.80M}       & \multicolumn{2}{l}{0.10W - 0.90B}      \\ \midrule
\multicolumn{1}{c|}{Method}                  & \multicolumn{1}{c}{FD $\downarrow$} & \multicolumn{1}{c|}{FID $\downarrow$} & \multicolumn{1}{c}{FD $\downarrow$} & \multicolumn{1}{c}{FID $\downarrow$} \\ \midrule
\multicolumn{1}{c|}{Random Sampling}                  & \multicolumn{1}{c}{0.478} & \multicolumn{1}{c|}{72.26} & \multicolumn{1}{c}{0.734} & \multicolumn{1}{c}{77.63} \\ \midrule
\multicolumn{1}{c|}{H-Distribution Sampling}                  & \multicolumn{1}{c}{\textbf{0.168}} & \multicolumn{1}{c|}{\textbf{51.65}} & \multicolumn{1}{c}{\textbf{0.325}} & \multicolumn{1}{c}{\textbf{53.80}} \\ \midrule
\end{tabular}
\end{adjustbox}
\label{tab:skew_bias_gen}
\end{table}

\noindent dataset. We present qualitative results for both the settings in Fig.\ref{fig:skew_generation}, where our distribution guidance is able to generate the defined distribution with majority \textit{males}. Note, we have binarized the race attribute as \textit{black} and \textit{whites} for simplicity, and hence in the generation, \textit{brown race} is considered under \textit{black} category. We report quantitative metrics in Tab.~\ref{tab:skew_bias_gen}, where the proposed method can achieve good FD scores with the same FID. Additional experimental results are tabulated in Sec.\textcolor{red}{B.1} of supplement.

\subsection{Ablations}
\vspace{-2mm} 
\label{subsec:ablate}
We provide ablation over the batch size here and that over the guidance scale, \textit{h-space} classifier architectures and number of training examples in Sec.\textcolor{red}{A} of supplement. 
\noindent \textbf{Batch size.} As we approximate $\mathbf{p^a_{\theta}}$ with a an estimate $\mathbf{\hat{p}^a_\theta}$ over a batch of size $N$, we ablate over different values of $N$ in Tab.~\ref{tab:ablate_bs}. Intuitively, using a larger batch size yields a better estimate of $\mathbf{p^a_{\theta}}$.
We have found that the $N=100$ works best for our experiments, balancing both FD and FID. Additionally, our model can also handle low data regime effectively.

\begin{table}[h]
\caption{Ablation over batch size for gender balancing.}
\vspace{-3mm}
\begin{adjustbox}{width=0.48\textwidth,center}
\begin{tabular}{@{}c|ccccccccccc@{}}
\toprule
Batch size & 2 & 4 & 8 & 10    & 25    & 50    & 75    & 100  & 125 & 150  & 200   \\ \midrule
FD   $\downarrow$    & 0.108 & 0.088 & 0.073  & 0.062 & 0.059 & 0.049 & 0.059 & \textbf{0.046} & 0.052 &  0.053 & 0.058 \\
FID   $\downarrow$   & 60.12 & 51.86 & 49.80  & 50.54 & 51.51 & 50.86 & 50.98 & 51.64 &  49.91 & 49.81 &  49.74 \\ \bottomrule
\end{tabular}
\end{adjustbox}
\label{tab:ablate_bs}
\end{table}


\begin{figure}[]
    \centering
    \includegraphics[width=\linewidth]{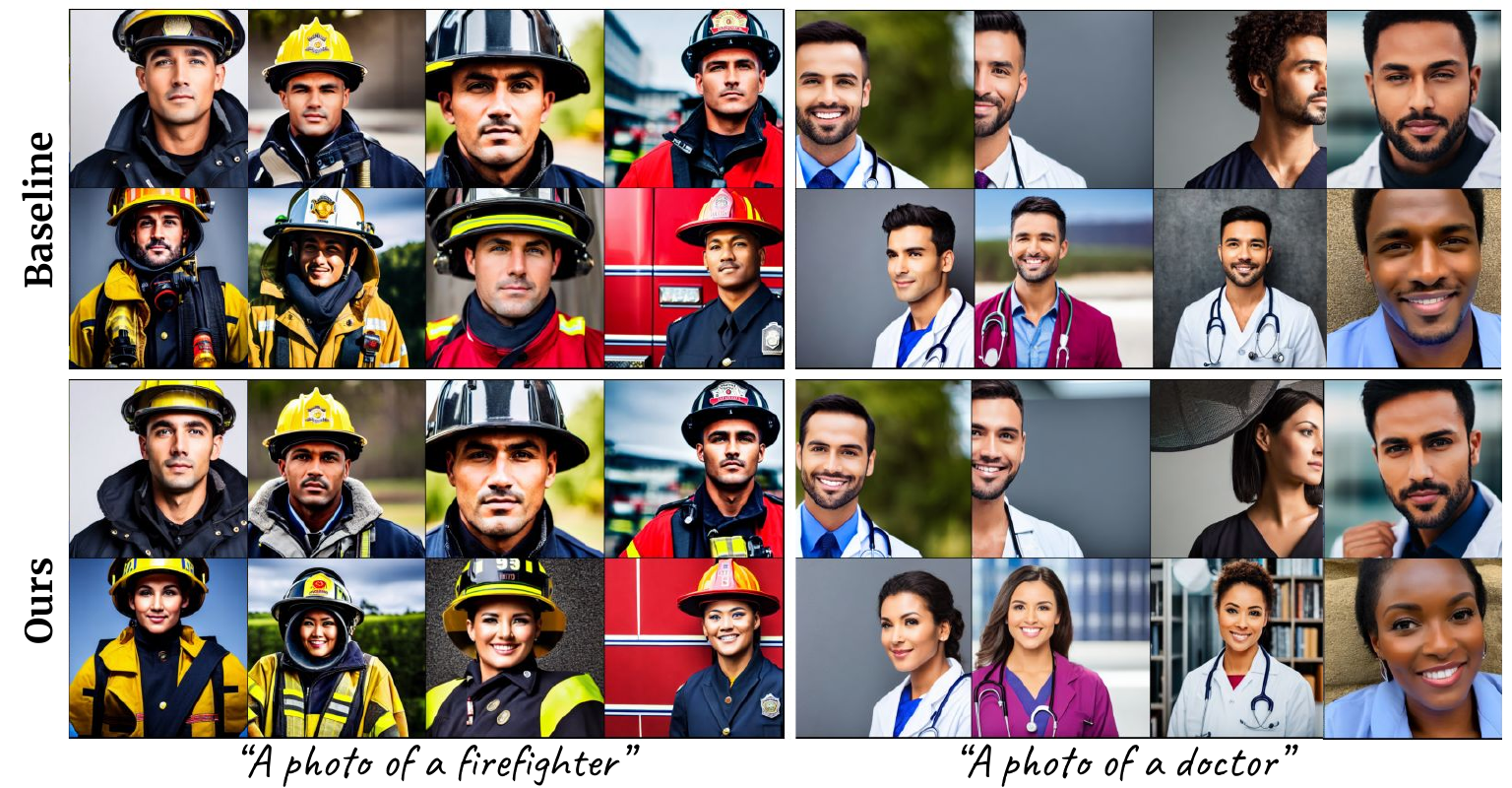} 
    \vspace{-8mm}
    \caption{Debiasing results on stable diffusion for gender. Distribution guidance can balance the \textit{gender} attributes in prompts involving other professions \eg. \textit{firefighter} and \textit{doctor}.}
    \label{fig:stable-diff}
    \vspace{-2mm} 
\end{figure}

\subsection{Debiasing Text-to-Image Diffusion Models} 
\label{subsec:sd_exp}
\vspace{-2mm}
We implemented our distribution guidance technique for debiasing a text-to-image generation model, Stable Diffusion (SD) v1.5~\cite{ldm} concerning the \textit{gender} attribute in this subsection. We provide results on other attributes and on mitigating spurious correlation on WaterBirds~\cite{waterbids_dataset} generation in Sec.\textcolor{red}{C} of supplement. First, we generate a dataset $\mathcal{D}$, of $10K$ images from SD, with prompts \textit{`a photo of a male'} and \textit{`a photo of a female'} to generate a labeled dataset for training \textit{h-space} classifiers. If the \textit{h-space} classifier is trained on CelebA-HQ dataset, guidance is ineffective due to a significant domain shift from the SD generations. Next, we obtain the corresponding labeled dataset $\mathcal{D}^h_{clf}$ in the \textit{h-space} and train a \textit{gender} classifier in the h-space\textit{h-space}. We used the trained classifier for distribution guidance following Sec.~\ref{subsec:guide_h_space} to generate images with balanced \textit{gender}.  

It is observed that SD increases the gender bias when queried to generate certain professions (\eg male and doctor)~\cite{zhang2023iti}. To this end, we implement our distribution guidance along the prompts \textit{`a photo of a doctor'} and \textit{`a photo of a firefighter'} to evaluate the effectiveness in this challenging setting. The qualitative results are shown in Fig.\ref{fig:stable-diff}, and the quantitative results are reported in Tab.~\ref{tab:sd_metrics}. 

\noindent \textbf{Baselines.} We compare our method with the following methods. \textbf{(1)} Random sampling from SD ~\cite{ldm}. \textbf{(2)} ITI-Gen ~\cite{zhang2023iti} that learns prompt embeddings for each category of the attribute given image reference sets of each category. It then appends these prompts during generation to produce balanced images. \textbf{(3)} Fair Diffusion ~\cite{friedrich2023fairdiffusion} that uses a lookup table to recognize the biased concept from the text input and adds scaled attribute expressions to the prompt. Note, these baselines are explicitly designed to debias text-conditioned diffusion model; however, our method can debias both conditional and unconditional diffusion models.


\begin{table}[]
\caption{Balancing \textit{gender} on Stable Diffusion~\cite{ldm} model.}
\vspace{-3mm}
\begin{adjustbox}{width=0.40\textwidth, center}
\begin{tabular}{@{}c|cc|cc|cc@{}}
\toprule
\multicolumn{1}{c|}{\multirow{2}{*}{Method}} & \multicolumn{2}{c|}{Gender}        & \multicolumn{2}{c|}{Doctor}        & \multicolumn{2}{c}{Firefighter}    \\ \cmidrule(l){2-7} 
\multicolumn{1}{c|}{}                        & \multicolumn{1}{c|}{FD $\downarrow$}    & FID $\downarrow$   & \multicolumn{1}{c|}{FD $\downarrow$}    & FID $\downarrow$  & \multicolumn{1}{c|}{FD $\downarrow$}    & FID $\downarrow$  \\ \midrule
Random Sampling                              & \multicolumn{1}{c|}{0.317} & 72.37 & \multicolumn{1}{c|}{0.355} & 70.11 & \multicolumn{1}{c|}{0.235} & 71.86 \\
ITI-Gen ~\cite{zhang2023iti}                 & \multicolumn{1}{c|}{0.049} & \textbf{64.79} & \multicolumn{1}{c|}{0.072} & \underline{67.81} & \multicolumn{1}{c|}{0.184} & 70.12 \\
Fair Diffusion ~\cite{friedrich2023fairdiffusion}  & \multicolumn{1}{c|}{0.227} & 71.22 & \multicolumn{1}{c|}{0.035} & 74.37 & \multicolumn{1}{c|}{\textbf{0.036}} & \textbf{68.33} \\
H-Sample Guidance (ours)                          & \multicolumn{1}{c|}{\underline{0.026}} & 70.96 & \multicolumn{1}{c|}{\underline{0.021}} & 68.43 & \multicolumn{1}{c|}{0.097} & 70.42 \\
H-Distribution Guidance (ours)                     & \multicolumn{1}{c|}{\textbf{0.024}} & \underline{70.69} & \multicolumn{1}{c|}{\textbf{0.015}} & \textbf{67.36} & \multicolumn{1}{c|}{\underline{0.093}} & \underline{69.41} \\ \bottomrule
\end{tabular}
\end{adjustbox}
\label{tab:sd_metrics}
\end{table}


\subsection{Class-imbalance in attribute classification}
\vspace{-2mm}
We explore an important downstream application of our proposed approach in balancing minority classes by augmenting the under-represented classes with generated data. Specifically, we train a race classifier (labels obtained from $\mathcal{C}_\mathcal{A}$) on the CelebA-HQ~\cite{celebA-hq} dataset. The race classifier is an ImageNet~\cite{deng2009imagenet}-pretrained ResNet-18~\cite{he2016deep} encoder, followed by 2 MLP layers and a classifier. We manually oversample the \textit{White}s and undersample the \textit{Black}s in the training dataset such that the imbalanced dataset consists of $10k$ samples of white people and $1k$ black people (we keep the genders balanced within a race class). Consequently, the model performs poorly on the minority class (i.e., \textit{Black}) due to under-representation. Next, we augment (class-balanced) the training data by generating samples whose distribution is inversely proportional to the class counts in the training set to increase images of minority classes using our distribution guidance approach. This adds 9k images of only \textit{Black} race, and the classifier trained on this balanced data performs significantly better (Tab. \ref{tab:downstream_classifier}). Even when gender is balanced in both the classes, we observe a significant disparity in the accuracies for Male and Female \textit{Black} samples in the vanilla classifier. However, our proposed method helps reduce the accuracy gap between Black males and Black females. This shows a potential application in generating `class' balanced datasets to train models for other downstream tasks, which can also mitigate bias.

\begin{table}[]
\caption{Group-wise accuracies of identical classifiers trained on existing and generated (balanced) data}
\vspace{-3mm}
\begin{adjustbox}{width=0.45\textwidth,center}
\begin{tabular}{@{}c|c|c|c|c@{}}
\toprule
                    & Black female & White female & Black male & White male \\ \midrule
Vanilla classifier  & 75.76        & 98.96      & 70.29        & 99.93      \\
Balanced classifier & 91.04        & 97.33      & 90.72        & 97.68      \\ \bottomrule
\end{tabular}
\end{adjustbox}
\label{tab:downstream_classifier}
\vspace{-7mm}
\end{table}

\vspace{-2mm}
\section{Discussion}
\vspace{-2mm} 

\textbf{Limitations.} 
Although our method performs guidance in the \textit{h-space}, which is efficient compared to the image space guidance, it still requires additional training of \textit{h-space} classifiers. Another limitation is reliance on accurate attribute classifiers to obtain labels for training \textit{h-space} classifiers. 

\noindent \textbf{Future works.} An important future work is extending distribution guidance beyond de-biasing for controlled generation and data augmentation. In the context of debiasing DMs, extending the proposed approach without needing an attribute classifier or labeled data. 

\vspace{-1mm}
\section{Conclusion}
\vspace{-2mm}
In this work, we aim to mitigate biases from pretrained diffusion models without retraining - given only a desired reference attribute distribution.
We propose a novel approach leveraging \textit{distribution guidance} that jointly guides a batch of images to follow the reference attribute distribution. The proposed method is effective and results in both high-quality and \textit{fair} generations across multiple attributes and outperforms sample guidance strategies based on conditioning each sample individually. Extensive experiments demonstrate the effectiveness of our method in balancing both single and multiple attributes on unconditional DMs and conditional text-to-image diffusion models. We believe such a setting of debiasing without retraining is practical, especially in today's era of large-scale generative models.  

\vspace{2mm }
\noindent \textbf{Acknowledgements.} 
This work was supported by the Kotak IISc AIML Centre (KIAC) and Meesho. Rishubh Parihar and Abhipsa Basu are supported by PMRF fellowship.  


{
    \small
    \bibliographystyle{ieeenat_fullname}
    \bibliography{main}
}

\clearpage 
\let\addcontentsline\origaddcontentsline 








\begin{center}
\appendix
\Large \textbf{Appendix}
\end{center}


\vspace{6mm} 
\tableofcontents


\section{Ablation Experiments} 
\label{sec:ablate} 
\subsection{Guidance Strength $\gamma$}
\label{sec:gamma_ablation} 
\begin{figure}[h]
    \centering 
    \includegraphics[width=\linewidth]{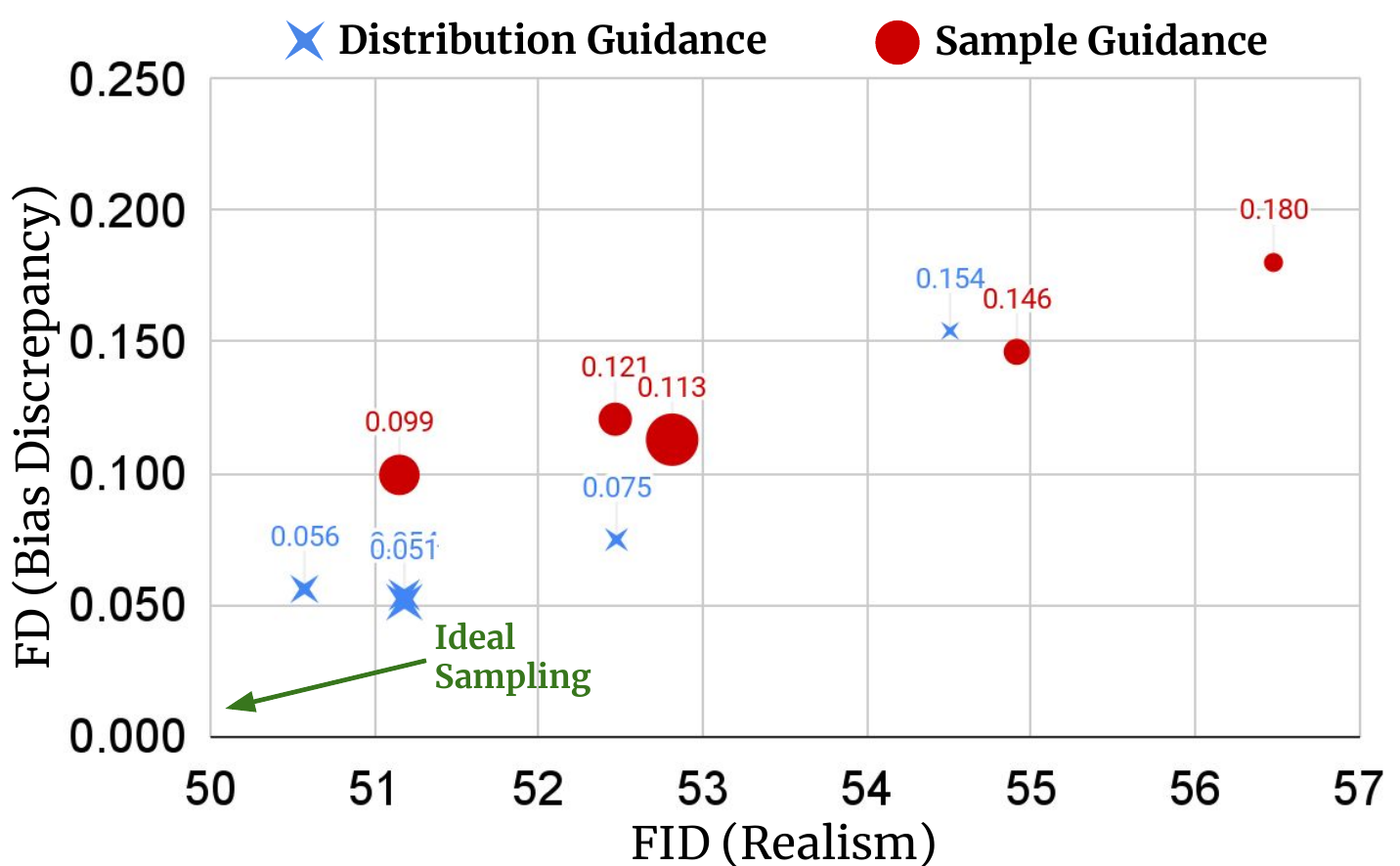}
    \caption{Ablation over guidance strength $\gamma$}
    \label{fig:guidance_ablation} 
\end{figure}

We ablate over guidance strength parameter $\gamma$ for both the distribution guidance and sample guidance in Fig.~\ref{fig:guidance_ablation}. Increasing $\gamma$ reduces the bias (better FD) at the cost of inferior image quality (increased FID). Guidance strength $1500$ achieved a good tradeoff between image quality and balancing attributes. Further, as compared to sample guidance, the distribution guidance achieves a better tradeoff for all the guidance strengths. 

\subsection{Data efficiency of h-space classifiers} 
\begin{figure}[h]
    \centering
    \includegraphics[width=\linewidth] {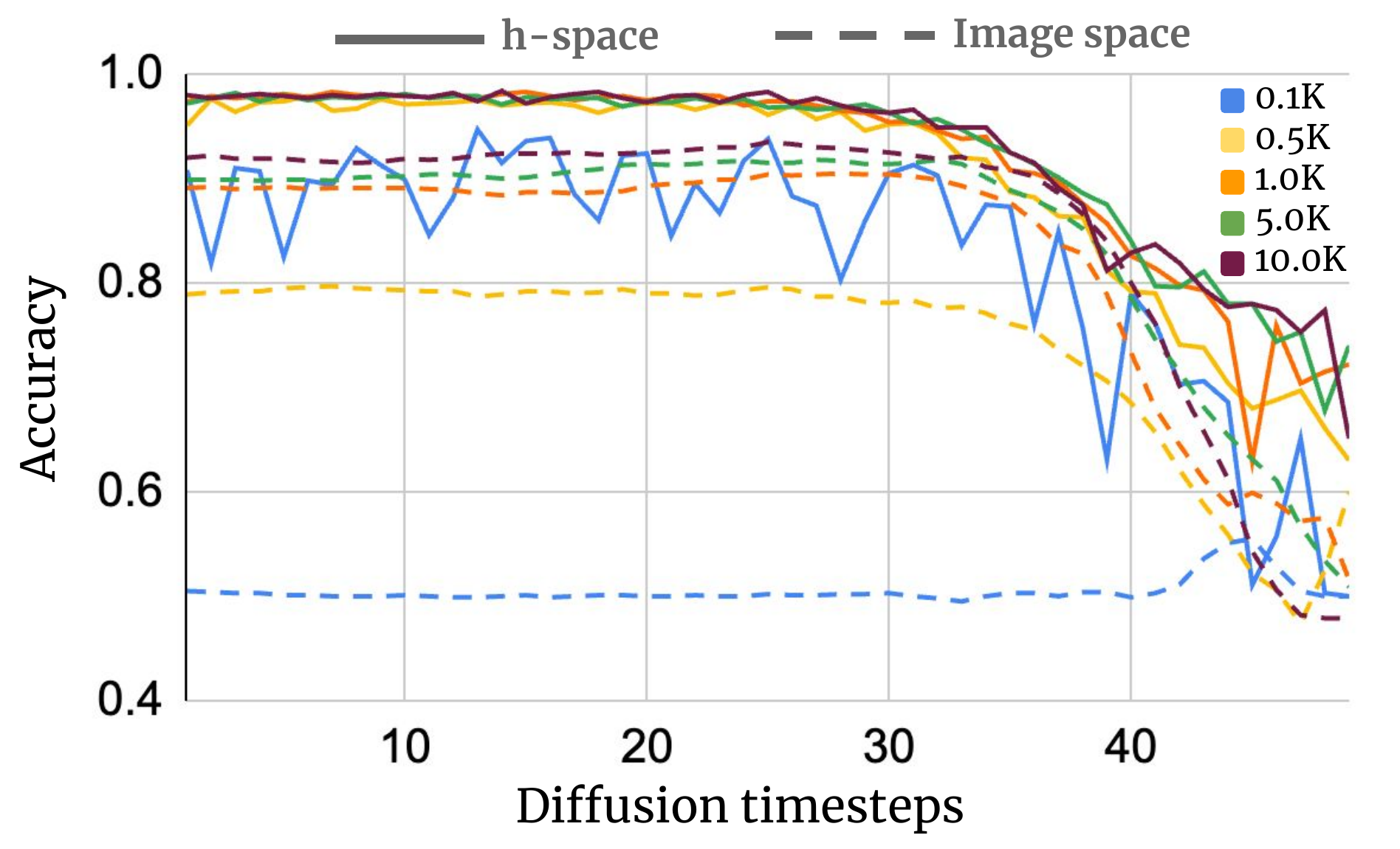}
    \caption{Ablation over the number of training examples for classifiers. h-space classifiers achieve excellent performance even with only $200$ training examples. In contrast, ResNet50 classifiers require a large number of images to achieve similar performance.}
    \label{fig:h-clfs-examples}
\end{figure}

We ablate over the number of training examples used to train the \textit{h-space} classifiers in Fig.~\ref{fig:h-clfs-examples}. Specifically, we train a \textit{gender}-attribute classifier in image space and h-space using the same number of images. We use ResNet50~\cite{he2016deep} as the image classifier and a linear head on top of the \textit{h-features} as the \textit{h-space} classifier. We pass the estimate $\hat{x}_0$ of the clean image at timestep $t$ to the image classifier to obtain the prediction. For \textit{h-space} classification, we pass the ${h_t}$ to the linear head. The \textit{h-space} classifiers are \textit{extremely data efficient} and can achieve $>90\%$ accuracy for most of the time-steps, even when trained with just $500$ training examples. This shows the efficacy of guidance in the h-space, which can be done with only a few hundred training examples. 

\subsection{H-space classifier architectures} We ablate over the different classifier architectures used for the classification in the h-space. We use a linear layer, an MLP and a small CNN architecture. The results are presented Tab.~\ref{tab:classifier_ab}. We report the average test accuracy over $50$ time steps, network parameters, and guidance time for generating a batch of images. With comparable accuracy, the linear layer has lesser parameters and performs guidance faster, and hence is used as the default classifier.

\begin{table}
\caption{Ablation over the classifier architecture for h-space classification}
\vspace{-3mm}
\label{tab:classifier_ab}
\begin{adjustbox}{width=0.9\linewidth,center}
\begin{tabular}{c|ccc}
\hline
Classifier & Avg. accuracy & Guidance time $\downarrow$ & \# Parameters \\ \hline
Linear & 0.921         & 1.0x          & 3.2M       \\
CNN    & 0.908         & 2.5x          & 74M        \\
MLP    & 0.952         & 2.0x          & 200M       \\ \hline
\end{tabular}
\end{adjustbox}
\end{table}



\section{Multi-attribute Debiasing}
\label{sec:expts}
In this section, we generate balanced subgroups for a combination of the multiple attributes. We consider jointly balancing $2$ and $3$ sensitive attributes across all subgroups. Specifically, we explore the following settings: \textbf{1) Gender + Race}: \textit{$0.25$ black males}, \textit{$0.25$ black females}, \textit{$0.25$ white males}, \textit{$0.25$ white females} \textbf{2) Gender + Eyeglasses}: \textit{$0.25$ males with eyeglasses}, \textit{$0.25$ females with eyeglasses}, \textit{$0.25$ males without eyeglasses} and \textit{$0.25$ females without eyeglasses}. \textbf{3) Gender + Race + Eyeglasses}: \textit{$0.125$ for all the $8$ subgroups formed}. 
The results are reported in Tab.~\ref{tab:subgroup-balance} where we compute the FD score with a balanced reference set and FID with the original CelebA-HQ~\cite{celebA-hq} dataset. The FID score quantifies the visual quality, whereas the FD score accounts for the bias in the generations, as explained in Sec.\textcolor{red}{4.1} (main paper). We do not use a reference set to compute FID as balancing across sub-groups leads to considerably less number of samples in the reference set.

\begin{table}[h]
\caption{Balancing attribute subgroups for fair generation}
\vspace{-3mm}
\label{tab:subgroup-balance}
\begin{adjustbox}{width=\linewidth}
\begin{tabular}{@{}c|cc|cc|cc@{}}
\toprule
                        & \multicolumn{2}{c|}{Gender + Race} & \multicolumn{2}{c|}{Gender + Eyeglasses} & \multicolumn{2}{c}{Gender + Race + Eyeglasses} \\ \midrule
Method                  & FD              & FID              & FD                 & FID                & FD                    & FID                    \\ \midrule
Random Generation       & 0.684                & 49.45                 & 0.636                  & 49.45                   & 0.768                & 49.45                       \\
Sample Guidance       & 0.436           & 45.49            & 0.3              & 47.41              &  0.496               & 47.83                       \\
Distribution Guidance & \textbf{0.224}  & 45.37            & \textbf{0.2}    & 45.92               &  \textbf{0.408}                &  43.94                      \\ \bottomrule 
\end{tabular}
\end{adjustbox} 
\end{table}

\begin{figure*}[h]
    \centering
    \includegraphics[width=\linewidth]{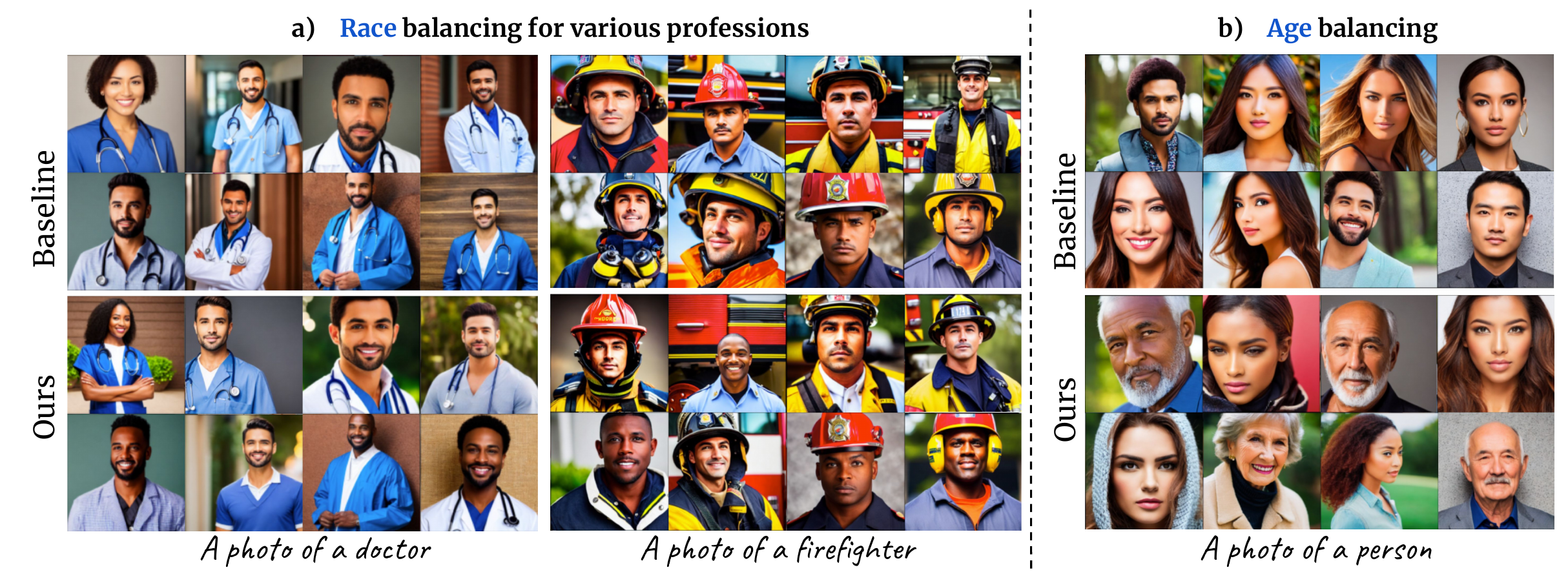}
    \caption{\textbf{Debiasing results on stable diffusion} for \textit{race} and \textit{age} attribute. We present random samples generated by the original stable diffusion model and with distribution guidance. \textbf{a)} Balancing across race involving different professions \eg firefighter, doctor. \textbf{b)} Balancing across age attribute for neutral prompts.}
    \label{fig:sd-face-debiasing}
\end{figure*}

\subsection{Generalization to imbalanced distributions}
\label{sec:imbalance}
Here, we present results for generating an imbalanced distribution for the subgroups by providing a skewed reference distribution. Such a setting is helpful in data augmentation for under-represented subgroups as shown in Sec.\textcolor{red}{4.8} in the main paper. We take the following two settings: \textbf{1) Gender + Eyeglasses}: \textit{$0.40$ males with eyeglasses, $0.10$ males without eyeglasses, $0.40$ females with eyeglasses, and $0.10$ females without eyeglasses}. \textbf{2) Gender + Race}: \textit{$0.40$ black males, $0.10$ white males, $0.40$ black females and $0.10$ white females}. These two configurations are contrary to the originally generated distribution as blacks and eyeglasses are minority groups. The results are provided in Tab.~\ref{tab:imbalance-distr}, where we report FD with a reference set and FID score with the CelebA-HQ dataset to evaluate bias and generation quality.  

\begin{table}[h]
\caption{Generating imbalance distribution across subgroups to generate more images for under-represented groups.}
\vspace{-3mm}
\label{tab:imbalance-distr}
\begin{adjustbox}{width=\linewidth}
\begin{tabular}{@{}c|cc|cc@{}}
\toprule
                        & \multicolumn{2}{c|}{Gender + Eyeglasses} & \multicolumn{2}{c}{Gender + Race} \\ 
                        & \multicolumn{2}{c|}{(0.40, 0.10, 0.40, 0.10)} & \multicolumn{2}{c}{(0.40, 0.10, 0.40, 0.10)} \\ 
                        \midrule
Method                  & FD                         & FID                         & FD                            & FID                            \\ \midrule
Random Generation       & 1.1                           & 49.45                            & 1.444                              & 49.45                               \\
Sample Guidance       & 0.472                      & 48.66                       &  0.756                        & 62.48                          \\
Distribution Guidance & \textbf{0.38}                      & 47.68                       &  \textbf{0.464}                        & 45.51                         \\ \bottomrule
\end{tabular}
\end{adjustbox} 
\end{table}

\section{Debiasing text-to-image Diffusion Model}
\label{sec:stable-diff-res}

\subsection{Social biases in facial attributes}
We extend our experiments with Stable diffusion (SD) ~\cite{ldm}, previously outlined in Sec. \textcolor{red}{4.6} (main paper) by implementing our distribution guidance technique for debiasing across other (\textit{race} and \textit{age}) attributes. We observe that SD generations with a neutral prompt, such as \textit{‘a photo of a firefighter’} or \textit{‘a photo of a doctor’}, are images of people who are predominantly of white origin. Similarly, when prompted with \textit{‘a photo a person’}, SD majorly generates images of young people. We attempt to mitigate these biases with the proposed distribution guidance method following Sec.\textcolor{red}{4.6}. The qualitative results for the same are provided in Fig.~\ref{fig:sd-face-debiasing}. As evident from the results, our method achieves fairness across \textit{race} and \textit{age} attributes using the same neutral prompts. We present quantitative results in Tab.~\ref{tab:sd-edits}, where our method achieves superior FD scores computed using CLIP attribute classifier as explained in Sec.\ref{sec:imple-details}.1.

\subsection{Background bias in bird generation}
\begin{figure}[h]
    \centering
    \includegraphics[width=0.9\linewidth]{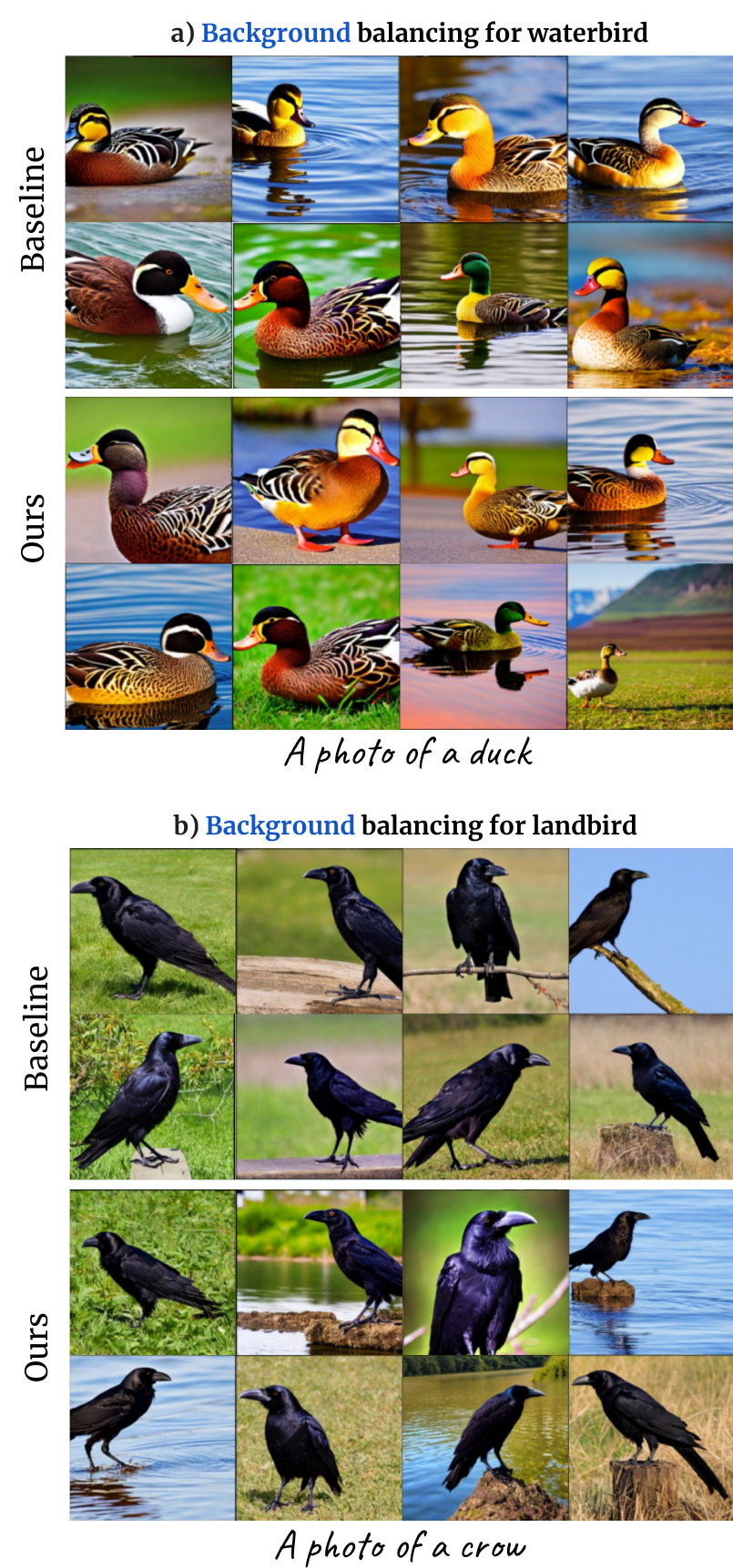} 
    \caption{\textbf{Debiasing results on stable diffusion for \textit{backgrounds} while generating birds}. The proposed \textbf{Distribution Guidance} can balance the number of birds in various backgrounds.}
    \label{fig:sd-bird-debiasing}
\end{figure}

Our method is applicable for mitigating biases that are not social as well (i.e. spurious correlations). WaterBirds~\cite{waterbids_dataset} is a widely used synthetic dataset for demonstrating spurious correlations, which consist of images of birds across different backgrounds. The images of birds (landbirds and waterbirds) are collated from the CUB dataset~\cite{cub_dataset}, and the backgrounds from the Places dataset~\cite{places_dataset}. The majority of the waterbirds (sea birds) are affixed with a water background, and landbirds with land backgrounds. We generate a dataset of $10K$ images using SD with prompts \textit{‘photo of a land background’} and \textit{‘photo of a water background’}. Additional negative prompts like - \textit{ocean, water, sea, shore, river} are added to generate pure \textit{land} images. Similarly negative prompts \textit{land, mountain, sand, forest, rocks} are added to generate pure \textit{water} images. Next, we train a background \textit{h-space} classifier for guidance. When queried with neutral prompts for landbird and waterbird, SD follows the spurious correlation in its generation. Specifically, when prompted for a landbird - \textit{‘a photo of a crow’}, SD predominantly generates a crow with land as the background, whereas when prompted for a waterbird - \textit{‘a photo of a duck’}, SD generates images with water in the background. When we apply distributional guidance, an equal number of images across both backgrounds are generated irrespective of the bird type. See Fig.~\ref{fig:sd-bird-debiasing} for qualitative results. 

\begin{table}[]
\caption{Balancing \textit{race} for multiple professions and \textit{age}}
\vspace{-3mm}
\centering
\label{tab:sd-edits}
\begin{adjustbox}{width=\linewidth}
    \begin{tabular}{c|ccc}
\hline
Method                       & Race-Doctor & Race-Firefighter & Age\\ \hline
Random Generation            & 0.356  & 0.423 &  0.488     \\
Distribution Guidance  & \textbf{0.191}  & \textbf{0.186}  & \textbf{0.194}     \\ \hline
\end{tabular}
\end{adjustbox}

\end{table}

\section{Implementation details}
\label{sec:imple-details}

\subsection{h-space classifiers}
\label{subsec:h-clf}
\textbf{Training data}. We created a paired training of \textit{h-space} features and attribute labels $\mathcal{D}^h_{clf}$ for training \textit{h-space} attribute classifiers. We start with CelebA-HQ~\cite{pro-gan} dataset $\mathcal{D}_{unf}$. Next, we used an off-the-shelf image space attribute classifier to obtain attribute labels for $\mathcal{D}$. We used CLIP as a classifier for both \textit{gender} and \textit{race} attributes as \textit{race} labels are not present in CelebA-HQ dataset. Specifically, we pass $\mathcal{D}_{unf}$ to the CLIP\cite{clip} image encoder and obtain its similarity with text prompts - \textit{`a male'} and \textit{`a black person'}. The highest and lowest similarity images are then filtered to create a labeled attribute dataset $\mathcal{D}$. We used $|\mathcal{D}| = 2000$, where $1000$ images are for the positive class, and others are for the negative class unless mentioned otherwise. For \textit{eyeglass} attribute, we used ground truth labels from the CelebA-HQ dataset, as the predictions from CLIP were inaccurate. We then embed $\mathcal{D}$ into the \textit{h-space} representation using DDIM\cite{ddim} inversion to obtain labeled dataset $\mathcal{D}^h_{clf}$ to train the classifiers.   

\noindent \textbf{Model architecture}. The h-space classifiers are implemented as a single linear layer for each diffusion timestep $\mathbf{t}$. We used DDIM inversion with $T=49$ timesteps and obtain $\mathcal{H}^i = \{h^i_t\}^{t=49}_{t=0}$ as a set of $49$ \textit{h-vectors} for each image $\mathbf{i}$. As each classifier is linear with two output neurons (positive/ negative class), they can be jointly represented as a single \textit{fully connected} layer with $2T$ output neurons. 

\noindent \textbf{Optimization}. We train the \textit{h-space} classifiers with the following hyper-parameters - batch size $64$, learning rate $0.001$, and for $5$ epochs on a single NVIDIA $A5000$ gpu. The overall training time for a single attribute classifier is 484.37s.  

\vspace{3mm}
\subsection{Evaluation Metrics}
We created a reference set $\mathcal{D}_{ref}$ of $5K$ images following the reference attribute distribution $\mathbf{p^a_{ref}}$, using the attribute labels from CelebA-HQ~\cite{celebA-hq}. The ground truth labels were used for gender and eyeglasses from CelebA-HQ. For race, however, as the labels are not available, they were obtained from CLIP as discussed in Sec.~\ref{subsec:h-clf}. We use $\mathcal{D}_{ref}$ to compute the reference statistics for FID. For FD, we compute the discrepancy of the predicted attribute distribution from the reference distribution following Sec.\textcolor{red}{4.1}. We use the Resnet-18~\cite{he2016deep} architecture to implement the attribute classifiers trained on CelebA-HQ~\cite{celebA-hq}. This set of classifiers \textit{needs} to be different from the one used to obtain $\mathcal{D}^h_{clf}$ for fair evaluation. 






\end{document}